%% file: 0_main_arxiv.tex
\PassOptionsToPackage{usenames,dvipsnames}{xcolor}
\PassOptionsToPackage{hyphens}{url}
\documentclass[nonacm,manuscript]{acmart}
\input{math_commands.tex}

\usepackage{graphicx}
\usepackage{soul}
\usepackage{multirow}
\usepackage{booktabs}
\usepackage{subcaption}
\usepackage{enumitem}
\usepackage{balance}
\usepackage{bbding}
\usepackage{colortbl}
\usepackage{longtable}
\AtBeginDocument{%
  }

\copyrightyear{2022}
\acmYear{2022}
\setcopyright{othergov}\acmConference[CIKM '22]{Proceedings of the 31st ACM
International Conference on Information and Knowledge Management}{October
17--21, 2022}{Atlanta, GA, USA}
\acmBooktitle{Proceedings of the 31st ACM International Conference on Information
and Knowledge Management (CIKM '22), October 17--21, 2022, Atlanta, GA, USA}
\acmPrice{15.00}
\acmDOI{10.1145/3511808.3557073}
\acmISBN{978-1-4503-9236-5/22/10}

\begin{document}

\title{Towards an Awareness of Time Series Anomaly Detection Models' Adversarial Vulnerability}

\author{Shahroz Tariq}
\affiliation{%
  \institution{
  Data61 CSIRO}
     \city{Sydney}
   \state{}
   \country{Australia}
}
\email{shahroz.tariq@csiro.au}

\author{Binh M. Le}
\affiliation{%
  \institution{College of Computing and Informatics, Sungkyunkwan University, South Korea}
  \city{}
  \state{}
  \country{}
}
\email{bmle@g.skku.edu}

\author{Simon S. Woo}
\authornote{Corresponding author \Envelope}

\affiliation{%
  \institution{
  Department of Artificial Intelligence, 
  Sungkyunkwan University, South Korea}
     \city{}
   \state{}
   \country{}
}
\email{swoo@g.skku.edu}

\begin{abstract}
  Time series anomaly detection is extensively studied in statistics, economics, and computer science. Over the years, numerous methods have been proposed for time series anomaly detection using deep learning-based methods. Many of these methods demonstrate state-of-the-art performance on benchmark datasets, giving the false impression that these systems are robust and deployable in many practical and industrial real-world scenarios. In this paper, we demonstrate that the performance of state-of-the-art anomaly detection methods is degraded substantially by adding only small adversarial perturbations to the sensor data. We use different scoring metrics such as prediction errors, anomaly, and classification scores over several public and private datasets ranging from aerospace applications, server machines, to cyber-physical systems in power plants. Under well-known adversarial attacks from Fast Gradient Sign Method (FGSM) and Projected Gradient Descent (PGD) methods, we demonstrate that state-of-the-art deep neural networks (DNNs) and graph neural networks (GNNs) methods, which claim to be robust against anomalies and have been possibly integrated in real-life systems, have their performance drop to as low as 0\%. To the best of our understanding, we demonstrate, for the first time, the vulnerabilities of anomaly detection systems against adversarial attacks. The overarching goal of this research is to raise awareness towards the adversarial vulnerabilities of time series anomaly detectors. 
\end{abstract}




\maketitle

\input{arxiv_sections/1_introduction}
\input{arxiv_sections/2_related}
\input{arxiv_sections/3_Methodology}
\input{arxiv_sections/4_Experiment}

\input{arxiv_sections/5_Results.tex}

\input{sections/6_Discussion}
\balance
\input{sections/7_Conclusion}


\bibliographystyle{ACM-Reference-Format}
\balance
\bibliography{0_main_arxiv}

\appendix
\input{arxiv_sections/8_Appendix}
\clearpage
\newpage

\end{document}

%% file: math_commands.tex
\usepackage{amsmath,amsfonts,bm}

\def\eqref#1{equation~\ref{#1}}

\def\1{\bm{1}}

\DeclareMathAlphabet{\mathsfit}{\encodingdefault}{\sfdefault}{m}{sl}
\SetMathAlphabet{\mathsfit}{bold}{\encodingdefault}{\sfdefault}{bx}{n}

\newcommand{\R}{\mathbb{R}}

\newcommand{\normlone}{L^1}
\newcommand{\normltwo}{L^2}

\newcommand{\normmax}{L^\infty}


%% file: arxiv_sections/1_introduction.tex
\section{Introduction}

\label{sec:intro}
Machine learning and deep learning have profoundly impacted numerous fields of research and society over the last decade~\citep{SAM1,goodfellow2016deep}. Medical imaging~\citep{MedicalImagingSurvey}, speech recognition~\citep{SpeechRecognitionSurvey}, environmental sciences~\cite{zeb1,zeb2} and smart manufacturing systems~\citep{SmartManufacturingSurvey} are a few of these areas. With the proliferation of smart sensors, massive advances in data collection and storage, and the ease with which data analytics and predictive modeling can be applied, multivariate time series data obtained from collections of sensors can be analyzed to identify 
particular patterns that can be interpreted and exploited. Numerous researchers have been interested in time series anomaly detection~\citep{AnomalySurvey,CLMPPCA,ITAD,YJ1,Shah_KDD_workshop,Anomaly_simon1,Anomaly_simon2,Anomaly_simon3}. For instance, time series anomaly detection methods are used in the aerospace industry for satellite health monitoring~\citep{CLMPPCA,ITAD,OmniAnomaly}. 
These deep neural network-based solutions outperform the competition on a variety of benchmark datasets. However, as deep learning became more prevalent, researchers began to investigate the vulnerability of deep neural networks, particularly to adversarial attacks. In the context of image recognition, an adversarial attack entails modifying an original image in such a way that the modifications are nearly imperceptible to the human eye~\citep{Fawaz14}. The modified image is referred to as an adversarial image, as it will be classified incorrectly by the neural network, whereas the original image will be classified correctly. One of the most well-known real-world attacks involves manipulating the image of a traffic sign in such a way that it is misinterpreted by an autonomous vehicle~\citep{Fawaz15}. The most common type of attack is gradient-based, in which the attacker modifies the image in the direction of the gradient of the loss function relative to the input image, thereby increasing the rate of misclassification~\citep{Fawaz14,FGSM,PGD}.

\begin{figure}[t]
\centering
    \begin{subfigure}[t]{0.49\linewidth}
        \centering
        \includegraphics[width=1\linewidth]{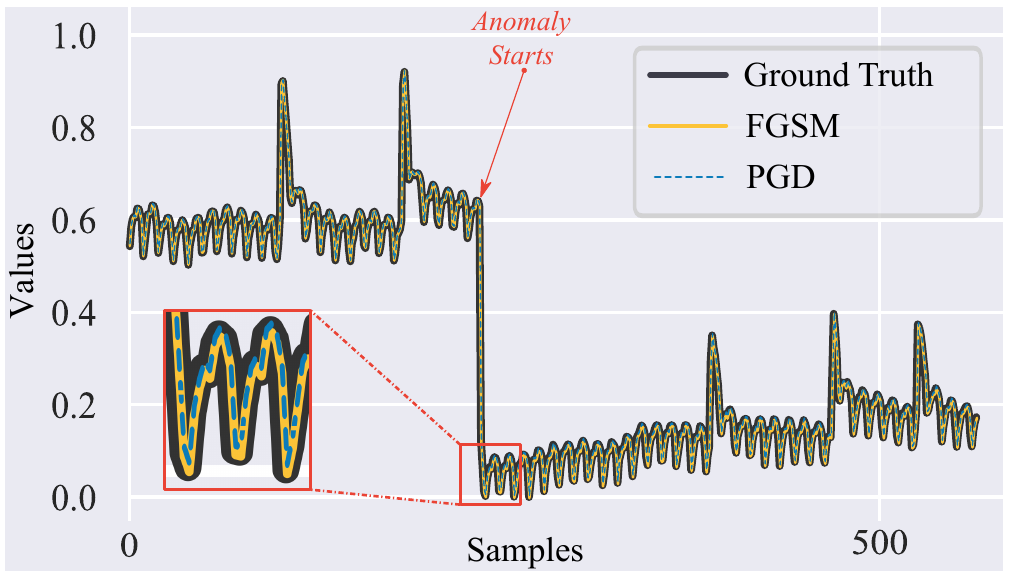}
    \end{subfigure}\hfill
    \begin{subfigure}[t]{0.49\linewidth}
        \centering
        \includegraphics[width=1\linewidth]{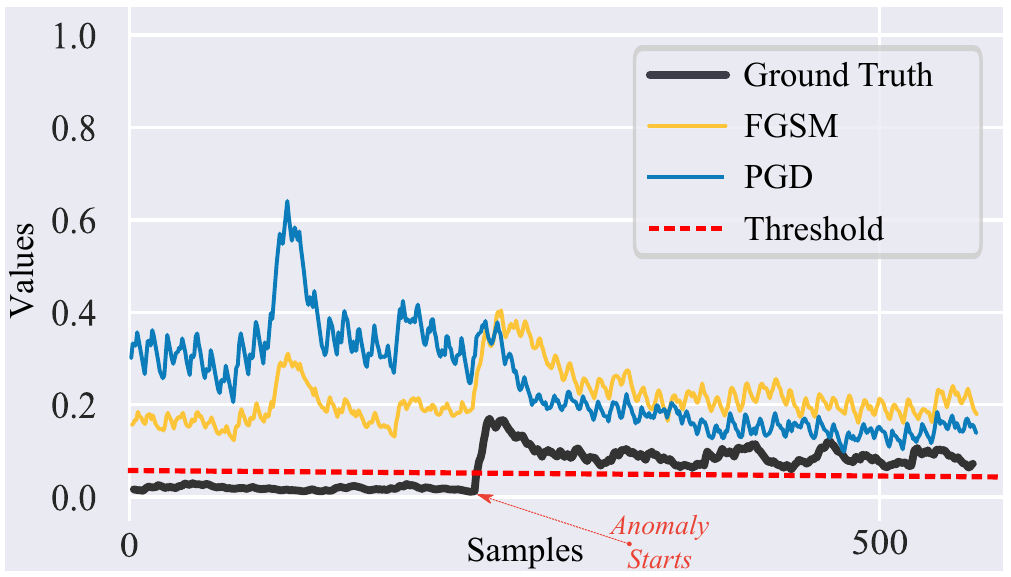}
    \end{subfigure}
    \caption{Example of ground truth and perturbed time series using FGSM and PGD attacks on CLMPPCA.}
  \label{fig:clmppca}
\end{figure}

While adversarial attacks have been extensively studied in the context of computer vision areas, they have not been extensively investigated for anomaly detection systems with time-series data. It is surprising to see much less research performed, given the increasing popularity of deep learning models for classifying time series~\citep{Fawaz2,Fawaz9,Fawaz19}. Additionally, adversarial attacks are possible in a large number of applications that require the use of time series data. 
For instance, Figure~\ref{fig:clmppca} (top) depicts the original and perturbed time series for the Korean Aerospace Research Institute's KOMPSAT-5 satellite (KARI)~\cite{CLMPPCA}. 
The prediction error (see Figure~\ref{fig:clmppca}, right) is generated by the Convolutional LSTM with Mixtures of Probabilistic Principal Component Analyzers (CLMPPCA) method~\cite{CLMPPCA}, which is currently incorporated at KARI, to predict anomalies. While CLMPPCA accurately predicts the anomaly for the original time series, adding small perturbations in the form of FGSM and PGD attacks causes the entire input samples to be classified as an anomaly. This attack can have a severe impact on the satellite health monitoring.

We present, transfer, and apply adversarial attacks that have been demonstrated to work well on images to time series data (containing anomalies) in this work. Additionally, we present an experimental study utilizing benchmark datasets from the aerospace and power plant industries and server machines, demonstrating that state-of-the-art anomaly detection methods are vulnerable to adversarial attacks. We highlight specific real-world use cases to emphasize the critical nature of such attacks in real-world scenarios. Our key findings indicate that deep networks for time series data, similar to their computer vision counterparts, are vulnerable to adversarial attacks. As a result, this paper emphasizes the importance of protecting against such attacks, particularly when anomaly detection systems are used in sensitive industries such as aerospace and power plants. 
Finally, we discuss some mechanisms for avoiding these attacks while strengthening the models' resistance to adversarial examples.

\textbf{Aim, Scope and Contribution. } 
In this work, we do not propose any novel adversarial attack method. However, we apply and demonstrate the threat of well-known existing adversarial attacks such as FGSM and PGD towards state-of-the-art anomaly detection methods for multivariate time-series data. In comparison to the computer vision domain, where adversarial attack has been extensively studied and investigated, the literature on novelty detection, and particularly on anomaly detection, is noticeably devoid of such studies. The purpose of this paper is to bring attention to this critical issue, whereas 
time series anomaly detection models also play pivotal roles in real-world scenarios as other vision tasks. Additionally, we hope to encourage researchers to consider robustness to adversarial attacks when evaluating time-series based future detectors. The paper's scope was limited to analyzing SOTA anomaly detectors. 
Finally, we successfully degraded the detection performance of deployed systems in the power plant and aerospace industries by employing adversarial attacks. It highlights how vulnerable the current generation of anomaly detectors is to adversarial attacks. Our source code and other implementation details are available here:   {\small \color{RoyalBlue}\url{https://github.com/shahroztariq/Adversarial-Attacks-on-Timeseries}}.












%% file: arxiv_sections/2_related.tex
\section{Related Work}
\label{sec:related}
In this section, we present background information, notations, and related works, with a particular emphasis on time series anomaly detection and adversarial attacks. 


\noindent
\textbf{Background and Notations. } When performing a supervised learning task, we define $D= \{(s_i,y_i)|i=1,\dots,N\}$ to represent a dataset containing N data samples. Each data sample is composed of a $m$-dimensional multivariate time series $s_i$ and a single target value $y_i$ for classification. However, the majority of anomaly detection occurs in an unsupervised setting. As a result, we take a slightly different approach from the supervised task. 
Hence, for unsupervised learning, each data sample is again composed of a $m$-dimensional multivariate time series $s_i$ however, $y_i$ is an $n$-dimensional multivariate time series obtained from an autoregressive model, predicting the future. In most cases, $n=m$ however, they can be different as well. Moreover, we define any deep learning method as
$\mathcal{F} (\cdot) \in \displaystyle f: \R^N \rightarrow \widehat{y}$
and loss function (e.g., cross entropy or mean squared error) as $\mathcal{L}_f(\cdot\,,\cdot)$. Finally, generating an adversarial instance $s_i^{adv}$ can be described as an optimization problem given a trained deep learning model $\mathcal{F}$ and an original input time series $s_i$, as follows:
\begin{equation}
\centering
\min \left \| s_i-s_i^{adv}\right \| 
  \;s.t.\;
  \mathcal{F}(s_i)  =\widehat{y_i},\;\;
  \mathcal{F}(s_i^{adv})=\widehat{y^{adv}_i} \;\; and \;\; \widehat{y_i} \ne \widehat{y^{adv}_i}
\end{equation}
\noindent
\textbf{Adversarial Attacks.}
In 2014, \citet{Fawaz24} introduced adversarial examples against deep neural networks for image recognition tasks for the first time. Following these inspiring discoveries, an enormous amount of research has been devoted to generating, understanding, and preventing adversarial attacks on deep neural networks~\citep{Fawaz15,FGSM,PGD}. Adversarial attacks can be broadly classified into two types: White-box and Black-box attacks. As White-box attacks presume access to the model's design and parameters, they can attack the model effectively and efficiently using gradient information. By contrast, Black-box attacks do not require access to the output probabilities or even the label, making them more practical in real-world situations. However, Black-box attacks frequently take hundreds, if not millions, of model queries to calculate a single adversarial case.

The majority of adversarial attack techniques have been proposed for use in image recognition. For instance, a Fast Gradient Sign Method attack was developed by~\citet{FGSM} as a substitute for expensive optimization techniques~\citep{Fawaz24}.
\citet{PGD} proposed Projected Gradient Descent (PGD) in response to the success of FGSM. PGD seeks to find the perturbation that maximizes a model's loss on a particular input over a specified number of iterations while keeping the perturbation's size below a specified value called epsilon ($\epsilon$). This constraint is typically expressed as the perturbation's $\normltwo$ or $\normmax$ norm. It is added to ensure that the content of the adversarial example is identical to that of the unperturbed sample — or even to ensure that the adversarial example is imperceptibly different from the unperturbed sample. Carlini-Wagner is another well-known attack \citep{CarliniWagnerL2}. However, it is primarily intended for $\normltwo$ norm-based attacks, whereas this study focuses exclusively on $\normmax$ norm-based attacks.

\noindent
\textbf{Adversarial Attacks on Time Series Anomaly Detectors. } 
Surprisingly, limited efforts have been made to extend computer vision-based adversarial attacks to time series anomaly detection domain. 
However, a few adversarial attack approaches have been proposed recently for the time series classification task, which are tangentially related to our work.
For instance, in their work on adopting a soft K Nearest Neighbors (KNN) classifier with Dynamic Time Warping (DTW), \citet{Fawaz26} demonstrated that adversarial examples could trick the proposed nearest neighbors classifier on a single simulated synthetic control dataset from the UCR archive~\citep{UCR}. Given that the KNN classifier is no longer considered the state-of-the-art classifier for time series data~\citep{fawaz27}, \citet{FawazADV} extend this work by examining the effect of adversarial attack on the more recent and commonly used ResNet classifier~\citep{ResNet}. \citet{FawazADV}, on the other hand, focused mainly on univariate datasets from the UCR repository. As a result, \citet{HarfordAdv} investigate the influence of adversarial attacks on multivariate time series classification using the multivariate dataset from UEA repository~\citep{UEA}. However, \citet{HarfordAdv} only consider basic methods such as 1-Nearest Neighbor Dynamic Time Warping~\citep{1NNDTW} (1-NN DTW) and a Fully Convolutional Network (FCN). \citet{fazle} and \citet{HarfordAdv} attacked models using Gradient Adversarial Transformation Networks (GATNs). 
However, they examined just transfer attacks, a relatively weak sort of Black-box attack. Only \citet{Siddiqui} demonstrated the effectiveness of gradient-based adversarial attacks on time series classification and regression networks. However, they considered a very simple baseline for the attack, containing only 3 convolutional, 2 max-pooling, and 1 dense layer.


\textbf{\textit{Note}: } Our study differs from previous research in that we focus on time series anomaly detection rather than the broader classification problem. More precisely, we explore autoregressive models that have been mostly overlooked in prior works. Additionally, rather than targeting generic deep neural networks KNN with DTW or ResNet, we investigate state-of-the-art anomaly detection methods. For instance, when it comes to anomaly detection, we focus on the most contemporary and commonly used techniques, such as MSCRED~\citep{MSCRED}, CLMPPCA~\cite{CLMPPCA}, and MTAD-GAT~\cite{MTADGAT}.
Section~\ref{sec:exp} will cover these methods in further depth.


%% file: arxiv_sections/3_Methodology.tex

\begin{figure*}
    \centering
    \includegraphics[
    width=1\linewidth]{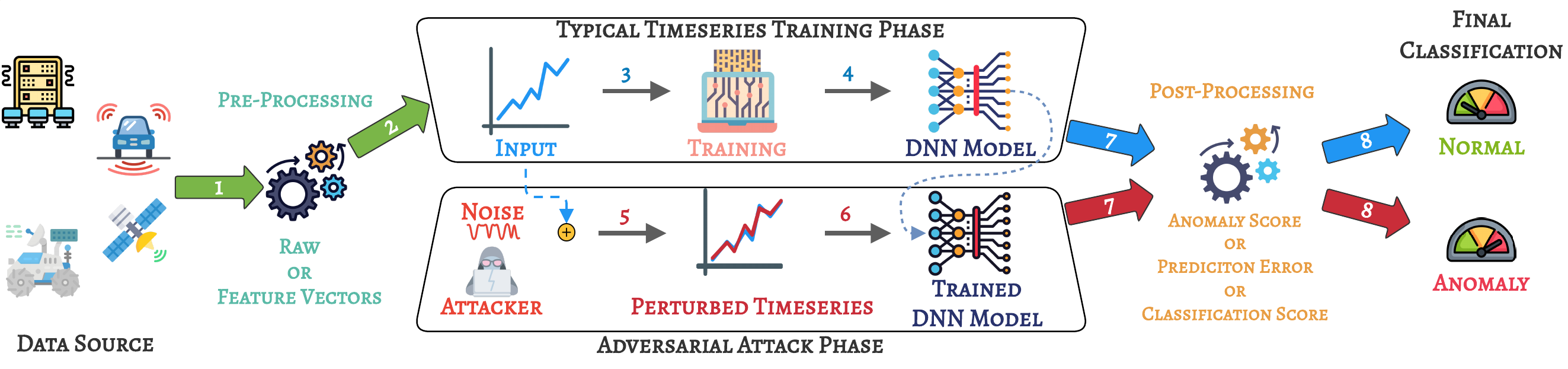}
    \caption{Pipeline of a typical time series anomaly detection's training phase and our  adversarial attack phase. }
    \label{fig:pipeline}
\end{figure*}

\section{Threat Model}
\label{sec:method}
To fully define the adversary, we divide the threat model into three subsections based on the adversary's capabilities, knowledge, and goals.

\noindent
\textbf{Adversary's Capabilities. }
We consider an adversary whose objective is to reduce the effectiveness of a victim model.  The attacker can apply the perturbations by modifying the victim's test-time samples, for example, by compromising a sensor or the data link that collects the data for inference. We investigate a $\normmax$ norm threat model with a $0.1$ epsilon. Due to the variable input range of time series data, there are no box constraints, in contrast to the visual image, where the pixels take on a definite value between $[0, 255]$. As a result, the data was standardized in our case using a zero-mean and unit standard deviation which justified the choice of $0.1$ as the epsilon value. 




\noindent
\textbf{Adversary's Knowledge. }
To evaluate the vulnerability of anomaly detection systems, we examine non-targeted White-box and Black-box scenarios. Typically, the attacker is given complete knowledge of the victim model, including its training data and the model's tunable parameters and weights.  However, we believe it to be unpractical in our scenario. As most of the system in our analysis are behind some layer of firewall or defense protection and most of the model parameters are hidden. Therefore, we consider two  types of adversary's knowledge as follows: 

\textbf{1) \textit{Complete Knowledge}: } The attacker understands how the model and its parameters works.  We can consider a \textit{White-box} attack to be the most appropriate method for this type of adversary. 

\textbf{2) \textit{Partial or No Knowledge}: } Given that the attacker has no or limited knowledge of the system, a \textit{Black-box} attack is the most appropriate method in this case. As a result, strategies such as transfer-based priors~\citep{TransferPrior} can be applied by the adversary.

\noindent
\textbf{Adversary's Goals. } 
The adversary considers two cases: (i) normal to anomaly and (ii) anomaly to normal. In (i), the adversary creates a $s_i^{adv}$ for each test sample $s_i$ so that the models interpret it as an anomaly, thereby generating a false-positive. However, in (ii), the adversary fabricates $s_i^{adv}$ to achieve the inverse effect, namely, to cause the model to predict an anomaly as normal, hence generating false-negative examples. As anomalies are rare events, even a few misclassifications caused by the adversary can have a detrimental effect on the model's performance.

\section{Adversarial Attack Generation}
The Fast Gradient Sign Method (FGSM) attack was proposed for the first time by \citet{FGSM}. The training of neural networks entails minimizing a loss function by adjusting the network weights. FGSM, on the other hand, does the opposite by adjusting the input samples in the direction opposite to the loss function's minimum. Thus, the FGSM attack is concerned with the computation of optimal perturbation series $\eta$, which can be added/summed to an input sample pointwise (i.e., a point refers to a single timestep) in order to maximize the classification loss function, i.e., cause misclassifications. This is mathematically expressed as:
\begin{equation}
    \eta = \epsilon \cdot \text{sign}\left (\nabla_s\;\mathcal{L}_f(s_i,y_i)\right )
\end{equation}
where $\nabla_s$ denotes the derivative of the network's loss, $\mathcal{L}_f(\cdot\,,\cdot)$, with respect to each timestep in $s_i$ (calculated for an input datapoint $s_i$ and it's true output $y_i$). To control the magnitude of the perturbation (i.e., to keep it imperceptibly small), $\epsilon$ is used as a multiplier factor. After that, the perturbed sample $s_i^{adv}$ can be computed as $s_i+\eta$. Note that FGSM requires the attacker to compute the loss function gradient with respect to a given input, which may not be possible directly. Due to the fact that FGSM requires knowledge of the internal workings of the network it is therefore referred to as a White-box attack. However, a surrogate model can be used to simulate the target model. An FGSM attack can be applied to the surrogate to generate adversarial examples~\citep{Mubarak15}, allowing for the use of such White-box attacks in practical scenarios~\citep{Mubarak16}.

\citet{PGD} proposed a more robust adversarial attack called Projected Gradient Descent (PGD). This attack employs a multi-step procedure and a negative loss function. It overcomes the problem of network overfitting and the shortcomings of the FGSM attack. It is more robust than first-order network information-based FGSM, and it performs well under large-scale constraints. Gradient Descent is essentially identical to the Basic Iterative Method (BIM)~\citep{Mubarak16} or the Iterative FGSM (IFGSM)~\citep{IFGSM} attacks. The only difference is that PGD initializes the example at a random location within the ball of interest (determined by the $\normmax$ norm) and performs random restarts, whereas BIM initializes at the original location.
\begin{equation}
\begin{split}
    s^{adv}_{i,t+1}=\Pi_{s+\delta}& \left (s^{adv}_{i,t} + \alpha\;\text{sign}(\nabla_s\mathcal{L}_f(s^{adv}_{i,t},y))\right)\\&\;\;\; s.t. \;\;\; 1\leq t \leq T
\end{split}
\end{equation}
\noindent
where $\delta$ is a nonempty compact topological space, $T$ is the total number of iterations,  and $\alpha$ is the control rate.
An illustration of the overall pipeline is provided in Figure~\ref{fig:pipeline}.


%% file: arxiv_sections/4_Experiment.tex
\section{Experimental Setup}
\label{sec:exp}
This section contains information on the benchmark datasets, evaluation metrics, criteria for selecting baselines, and chosen baselines.

\noindent
\textbf{Datasets} For anomaly detection we employ three public datasets: (i) Mars Science Laboratory rover (MSL) \citep{SMAPMSL}, (ii) Soil Moisture Active Passive satellite (SMAP) \citep{SMAPMSL}, and (iii) Server Machine Dataset (SMD) \citep{OmniAnomaly}, as well as one private dataset: (vi) Korean Aerospace Research Institute KOMPSAT-5 satellite (KARI) \citep{CLMPPCA} and one synthetic dataset: (v) from the MSCRED paper \citep{MSCRED}.
The datasets were chosen based on our baselines' shown ability to provide state-of-the-art performance on them. 
Table~\ref{tab:datasets} summarize these datasets.

\noindent
\textbf{Evaluation Metrics} 
To obtain the final classification result for anomaly detection methods, we observed that the majority of detectors use a thresholding method on top of the neural network's predictions, which are expressed as an anomaly score or prediction error. The precision, recall, and F1-score are then calculated using the results from thresholding methods. While these metrics are beneficial, the true impact of the adversarial attack is visible primarily in anomaly detectors' anomaly score and prediction errors. Therefore, we include Figure~\ref{fig:clmppca}, \ref{fig:mtad-gat} and \ref{fig:mscred}, as illustrations of this impact. Additionally, we include more related figures in Appendix~\ref{app_sec:MSCRED}--~\ref{app_sec:CLMPPCA}.
\begin{table}[t!]
\centering
\caption{A summary of anomaly detection datasets.}
\label{tab:datasets}
\begin{tabular}{lcccccc} 
\toprule
\textbf{Statistics} & \textbf{SMAP} & \textbf{MSL} & \textbf{SMD} & \textbf{KARI} & \textbf{Synthetic}\\ 
\hline
Dimensions & 55 & 27 & 28 & 4-35 & 30\\
Anomalies & 13.13\% & 10.27\% & 4.16\% & 1.00\% & 1.10\% \\
Train Size & 135,183 & 58,317 & 708,405 & 4,405,636 & 8,000 \\
Test Size & 427,617 & 73,729 & 708,420 & 17,622,546 & 10,000 \\ 
\bottomrule
\end{tabular}
\end{table}


\subsection{Anomaly Detection Baselines}
\subsubsection{\textbf{Selection Criteria}}
We conduct experiments on the following baselines to demonstrate that the vulnerability to adversarial attacks is common among several state-of-the-art anomaly detection architectures.
    Anomaly detectors based on Deep Neural Networks (DNNs) are the most frequently used method. However, some methods based on Graph Neural Networks (GNNs) have also been proposed recently. As a result, we evaluated both DNNs- and GNNs-based anomaly detectors. We used the following criteria to determine the baseline: 
    \begin{enumerate}
        \item \textbf{\textit{Diverse Architecture}: } To ensure that we cover a broad range of methods, we decide that the baselines should be diverse, i.e., no two baselines have similar model architecture.
        \item \textbf{\textit{Diverse pre-processing}: } They should consider a different pre-processing technique (e.g., using raw data or feature vectors). 
        \item \textbf{\textit{Diverse post-processing}: } They should take into account various post-processing techniques for prediction (e.g., anomaly score, prediction error, or classification score). 
        \item \textbf{\textit{Peer-reviewed}: } The method is widely accepted and peer-reviewed. For this criterion, we take into account GitHub Forks, paper citations, and publication venues.
        \item \textbf{\textit{Open-source}: } The source code is freely available or can be obtained upon request.
    \end{enumerate}

    \subsubsection{\textbf{Selected Baselines}}
    We choose the following baselines based on the aforementioned criteria:
    
    \textbf{\textit{MSCRED} \citep{MSCRED} [\textit{AAAI'19}]: } Taking advantage of the temporal dependencies inherent in multivariate time series, \citet{MSCRED} proposed a Multi-Scale Convolutional Recurrent Encoder-Decoder (MSCRED) for anomaly detection on two datasets: (i) synthetic and (ii) power plant. Sidenote: \citet{OneClassNeurIPS} demonstrated that MSCRED outperforms all SOTA anomaly detection methods except Temporal Hierarchical One-Class (THOC), but we were unable to evaluate THOC as the code is not available (see more details below this list). As a result, we chose the second best method (i.e., MSCRED) among recently developed SOTA anomaly detection methods. Because the power plant dataset is not publicly available, we compare MSCRED with and without adversarial attack using the synthetic dataset used by \citet{MSCRED} in their work. 
    
    \textbf{\textit{CLMPPCA} \citep{CLMPPCA} [\textit{KDD'19}]: } \citet{CLMPPCA} proposed a hybrid approach for anomaly detection in multivariate satellite telemetry data. Based on the accomplishments of Convolutional LSTM-based networks in understanding spatiotemporal data in various domains~\cite{CONVLSTM_s1,CONVLSTM_s2,CONVLSTM_s4,CONVLSTM_s3}, they propose a Convolutional LSTM with Mixtures of Probabilistic Principal Component Analyzers (CLMPPCA) method for transforming the time window containing several telemetry data samples into a feature vector that is used to train the model and to predict the future data instances. To make final classification, the prediction errors calculated from the prediction and ground truth are combined with a moving average-based threshold method. In their work \citet{CLMPPCA} used a private dataset from the Korean Aerospace Research Institute's (KARI) KOMPSAT-5 satellite for evaluation. We were able to obtain the same private dataset and demonstrate how adversarial attacks affect the performance of CLMPPCA. One of the primary reasons for selecting CLMPPCA is that it is currently deployed at KARI. Thus, successfully demonstrating an attack on this method will demonstrate its applicability in a practical scenario.
    
    \textbf{\textit{MTAD-GAT} \citep{MTADGAT} [\textit{ICDM'20}]: } \citet{MTADGAT} proposed a multivariate time series anomaly detector based on Graph Attention Networks. The authors treat each univariate time series as a separate feature and employ two parallel graph attention layers to learn the complex dependencies between multivariate time series in both temporal and feature dimensions by jointly optimizing a forecasting-based and reconstruction-based model. MTAD-GAT outperformed several recent time series anomaly detectors such as OmniAnomaly~\citep{OmniAnomaly}, MAD-GAN~\citep{MADGAN}, and DAGMM~\citep{DAGMM} from ICLR 2018, on three publicly available anomaly datasets (SMAP, MSL, and SMD). As a result, MTAD-GAT is one of the best SOTA methods currently available. We evaluate MTAD-GAT with and without adversarial attacks on all three datasets (i.e., SMAP, MSL, and SMD). 

\textbf{\textit{Note}:} We chose these three baselines based on their compliance with our defined criteria. Additionally, we were unable to evaluate some recent methods, such as Temporal Hierarchical One-Class (THOC) published at NeurIPS 2020 because the source code is not publicly available and our request to obtain the source code from the author was not answered. We discuss this further in Section~\ref{sec:repro}.

\subsubsection{\textbf{White- and Black-box Attack Settings.} }
As the attacker will have complete knowledge of the underlying system in a White-box attack, we create attack vectors using the same selected baselines, namely MSCRED, CLMPPCA, and MTAD-GAT. Whereas for the Black-box attack, we build attack vectors using a model that is similar to but simpler than the victim model. For example, we utilise a vanilla recurrent autoencoder to create attack vectors for MSCRED, a simple CNN+LSTM model for CLMPPCA, and a vanilla GNN for MTAD-GAT.

%% file: arxiv_sections/5_Results.tex
\section{Empirical Evaluation}
\label{sec:results}

We present results for the $\normmax$ FGSM and PGD attacks against three SOTA anomaly detection methods—MSCRED, CLMPPCA, and MTAD-GAT. 
The Appendix includes additional details about the $\normmax$, $\normlone$, and $\normltwo$ attacks results (Appendix~\ref{app_sec:MSCRED}); 
more details on impact of adversarial attacks on MTAD-GAT (Appendix~\ref{app_sec:MTAD-GAT}); 
some original vs. perturbed time series samples (Appendix~\ref{app_sec:CLMPPCA}. Moreover, results from the FGSM, PGD, BIM, Carlini-Wagner, and Momentum Iterative Method (MIM)~\citep{MIM} attacks on 71 datasets from the UCR repository are available on our GitHub repository. In general, we observe that perturbations that are $\normmax$-bounded are more effective. This could be explained by optimization challenges, as $\normlone$ and $\normltwo$ attacks are typically more difficult to optimize~\citep{CarliniWagnerL2,SL1D}.

\subsection{Adversarial Attack on MSCRED}
\subsubsection{\textbf{MSCRED (White-box)}}

We employ non-targeted FGSM and PGD methods to attack MSCRED. As a result, only $s_i$  from the test set is made available to the attack methods. The $\epsilon$ is set to $0.1$ for the FGSM attack, and $\alpha$ is set to $0.1$ for the PGD attack with $T=40$. The MSCRED method determines the appropriate threshold between normal and anomalous data points based on the training data. As a result, any modification to the test samples should not affect the threshold. As shown in Table~\ref{tab:mscred}, the victim model (MSCRED) has no efficacy on the samples perturbed by FGSM and PGD attacks and thus fails to detect all anomalies. Additionally, MSCRED classifies all instances of normal data as anomalies. We demonstrate in Figure~\ref{fig:mscred} that MSCRED (No Attack) can accurately predict the majority of anomalies with an F$_1$ score of $0.890$ (see Table~\ref{tab:mscred}). According to Figure~\ref{fig:mscred}, the anomaly scores under FGSM (yellow) and PGD (blue) attacks are always higher than the threshold (red dashed line), which means that MSCRED is predicting everything as an anomaly, resulting in an F$_1$ score of less than $0.50$.  It is intriguing that such a small amount of change in the time series, which is primarily imperceptible to the naked eye, can greatly affect the MSCRED's anomaly scores, even when the perturbations are so minute. The results in Table~\ref{tab:mscred} are demonstrating that MSCRED is not robust against adversarial attacks. 

\subsubsection{\textbf{MSCRED (Black-box)}} As with the White-box attack, the Black-box attack significantly reduced MSCRED's F1-score. This reduction, however, is slightly less than that caused by a White-box attack. Moreover, the PGD attack reduced F1-scores more than the FGSM attack, as shown in Table~\ref{tab:mscred}. This experiment demonstrates that even when we build the attack vector using a different backbone model, we can still achieve significant success by transferring the adversarial attack.

\subsection{Adversarial Attack on MTAD-GAT}
\subsubsection{\textbf{MTAD-GAT (White-box)}} As with MSCRED, we attack MTAD-GAT using a non-targeted FGSM and PGD method with $\epsilon=0.1$, $\alpha=0.1$, and $t = 40$. The results of adversarial attacks against MTAD-GAT trained on the MSL, SMAP, and SMD datasets are shown in Table~\ref{tab:mtad-gat}. MTAD-GAT demonstrates state-of-the-art performance for anomaly detection in the absence of an adversarial attack (No Attack). However, when adversarial examples from FGSM and PGD are used to evaluate it, the detection performance drops to as low as 66\%. The impact of the PGD attack is more significant than that of the FGSM attack, which is understandable given that PGD is a more powerful attack than FGSM. It leads us to ponder that if more sophisticated attacks are explicitly developed for time series data, they will have a significantly greater impact on SOTA anomaly detectors. As a result, future anomaly detection methods should take adversarial examples into account.



\begin{table}
\centering
\caption{MSCRED results (F$_1$ score) on synthetic dataset from original paper. Both White- and Black-box attacks show significant success with FGSM and PGD. We used surrogate model (vanilla recurrent autoencoder) to generate adversarial examples for Black-box attack.}
\label{tab:mscred}
\begin{tabular}{l|ccc|ccc} 
\toprule
\multirow{3}{*}{\textbf{Method}} & \multicolumn{3}{c|}{\textbf{White-box }} & \multicolumn{3}{c}{\textbf{Black-box }} \\
 & \textit{\textbf{Pre.}} & \textit{\textbf{Rec.}} & \textit{\textbf{F$_1$}} & \textit{\textbf{Pre.}} & \textit{\textbf{Rec.}} & \textit{\textbf{F$_1$}} \\ 
\cline{2-7}
No Attack & 1.000 & 0.800 & 0.890  & 1.000 & 0.800 & 0.890  \\
FGSM & 0.487 & 0.500 & 0.493 & 0.651 & 0.693 & 0.671  \\
PGD & \textbf{0.485} & \textbf{0.500} & \textbf{0.492} & \textbf{0.634} & \textbf{0.677} & \textbf{0.655} \\
\bottomrule
\end{tabular}
\end{table}



\begin{table}
\centering
\caption{MTAD-GAT results (F$_1$ score) on MSL, SMAP and SMD datasets. For all three datasets, both White- and Black-box attacks are highly effective. We used vanilla GNN  as a surrogate model to generate adversarial examples for Black-box attack.}
\label{tab:mtad-gat}
\begin{tabular}{l|ccc|ccc} 
\toprule
\multirow{3}{*}{\textbf{Method}} & \multicolumn{3}{c|}{\textbf{White-box}} & \multicolumn{3}{c}{\textbf{Black-box}} \\
 & \textbf{\textit{MSL}} & \textbf{\textit{SMAP}} & \textbf{\textit{SMD}} & \multicolumn{1}{c}{\textbf{\textit{MSL}}} & \multicolumn{1}{c}{\textbf{\textit{SMAP}}} & \multicolumn{1}{c}{\textbf{\textit{SMD}}} \\ 
\hline
No Attack & 0.950 & 0.894 & 0.999 & 0.950 & 0.894 & 0.999 \\
FGSM & 0.719 & 0.804 & 0.803 & 0.751 & 0.847 & 0.852 \\
PGD & \textbf{0.687} & \textbf{0.775} & \textbf{0.665} & \textbf{0.727} & \textbf{0.815} & \textbf{0.749} \\
\bottomrule
\end{tabular}
\end{table}

\begin{figure}[t!]
\centering
    \begin{subfigure}[t]{0.5\linewidth}
        \centering
        \includegraphics[clip, trim=0pt 0pt 0pt 0pt,width=0.8\textwidth]{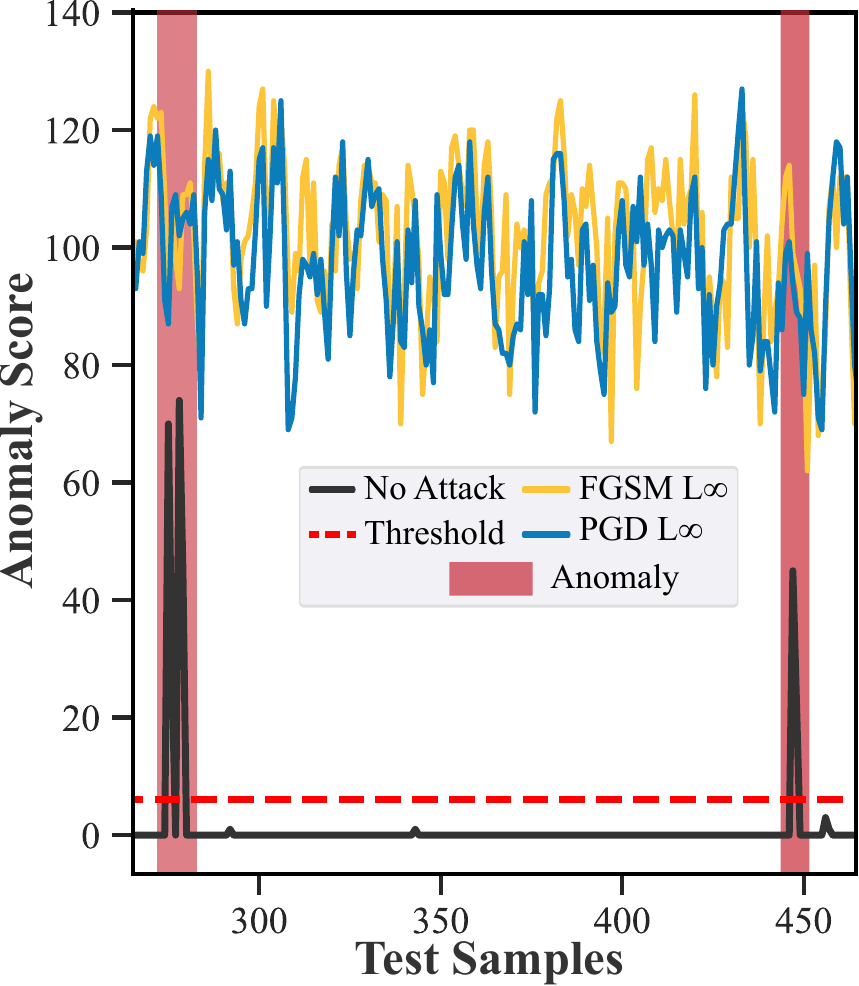}
        \caption{No Attack, FGSM and PGD on MSCRED}
        \label{fig:mscred}
    \end{subfigure}\hfill
    \begin{subfigure}[t]{0.5\linewidth}
        \centering
        \includegraphics[clip, trim=0pt 0pt 0pt 0pt,width=1\linewidth]{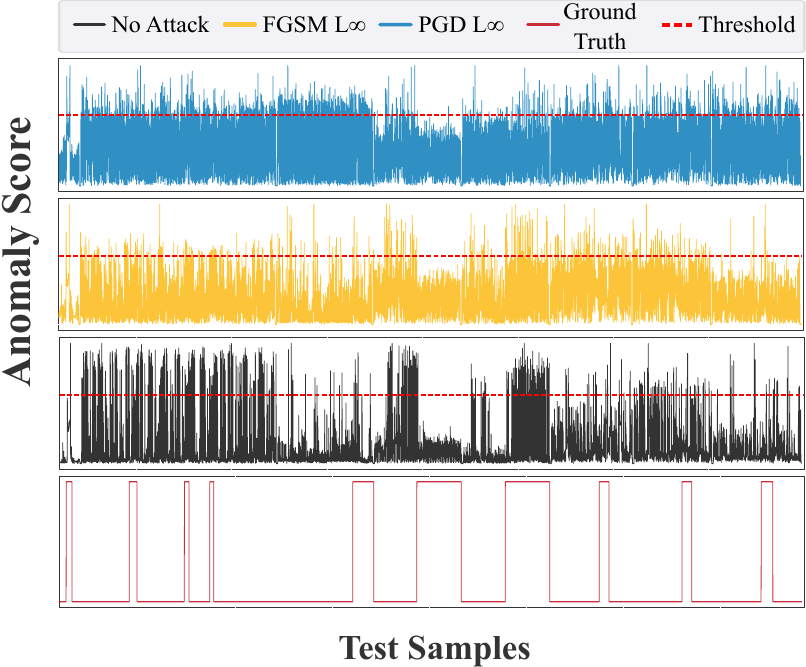}
        \caption{No Attack, FGSM and PGD on MTAD-GAT}
        \label{fig:mtad-gat}
    \end{subfigure}\hfill
    \caption{Anomaly score of No Attack, FGSM and PGD on MSCRED (a) and on MTAD-GAT for MSL dataset (b) The y-axis scale is between 0 and 1 for (b). See Appendix~\ref{app_sec:MSCRED} and ~\ref{app_sec:MTAD-GAT} for more details on (a) and (b), respectively.} 
    \label{fig:Results}
\end{figure}



    

Additionally, Figure~\ref{fig:mtad-gat} illustrates the effect of adversarial examples from FGSM (yellow) and PGD (blue) attacks on the MTAD-GAT anomaly score for the MSL dataset.  We can see that the anomaly scores for FGSM and PGD frequently exceed the threshold (red dashed line), resulting in a large number of false positives and lowering the F$_1$ score from 94.98\% to 71.90\% for FGSM and 68.69\% for PGD.

\subsubsection{\textbf{MTAD-GAT (Black-box)}} Like the Black-box attack on MSCRED, the attack on MTAD-GAT has a similar effect, lowering the F1-score for MSL, SMAP, and SMD to 0.751, 0.847, and 0.852 with the FGSM attack and 0.727, 0.815, and 0.749 with the PGD attack, as shown in Table~\ref{tab:mtad-gat}. As with the Autoencoder, this experiment indicates that it is possible to transfer adversarial attack to graph neural networks. Thus, demonstrating that the adversary may not require extensive knowledge of the backbone to launch a successful attack.

\subsection{Adversarial Attack on CLMPPCA}
\subsubsection{\textbf{CLMPPCA (White-box)}} The KARI dataset is divided into ten subsystems. As a result, we trained the CLMPPCA model on each subsystem separately, as described in the original paper. We then used FGSM and PGD attacks to evaluate each of these trained models. For FGSM, we use $\epsilon=0.1$, for PGD, we use $\alpha=0.1$, and $t = 40$. Table~\ref{tab:clmppca} summarizes the prediction errors for each subsystem prior to and following the attack. We can see that when adversarial attacks are used, the prediction error increases up to twentyfold. Note: For brevity and space constraints, we omit the F$_1$ score from Table~\ref{tab:clmppca}, as it is $0.50$ for all subsystems. CLMPPCA fails to predict any anomalies under FGSM and PGD attacks because the prediction error is always higher than the threshold (see Figure~\ref{fig:clmppca}). We believe that by employing these straightforward yet effective attacks, an adversary can easily introduce false positives into CLMPPCA's predictions at will, posing significant difficulties for satellite operators.

\begin{table*}
\centering
\caption{CLMPPCA prediction errors for subsystems (SS) 1-10 on the KARI KOMPSAT-5 dataset from FGSM and PGD attack. The prediction errors enclosed in brackets are the result of Black-box attack, whereas those outside the brackets are from White-box attack. A higher error value indicates a more powerful attack.}
\label{tab:clmppca}
\begin{tabular}{l|cccccccccc} 
\toprule
\textbf{Methods} & \textbf{SS1} & \textbf{\textbf{SS}2} & \textbf{\textbf{SS}3} & \textbf{\textbf{SS}4} & \textbf{\textbf{SS}5} & \textbf{\textbf{SS}6} & \textbf{\textbf{SS}7} & \textbf{\textbf{SS}8} & \textbf{\textbf{SS}9} & \textbf{\textbf{SS}10} \\ 
\hline
No Attack & 0.025 & 0.020 & 0.646 & 0.018 & 0.078 & 0.081 & 0.028 & 0.015 & 0.043 & 0.106 \\ 
\hline
\multirow{2}{*}{FGSM } & 0.306 & 0.327 & 5.657 & 0.153 & 1.744 & 1.708 & 0.246 & 0.201 & 1.303 & 0.314 \\
 & \multicolumn{1}{l}{(0.132)} & \multicolumn{1}{l}{(0.159)} & \multicolumn{1}{l}{(3.163)} & \multicolumn{1}{l}{(0.092)} & \multicolumn{1}{l}{(0.680)} & \multicolumn{1}{l}{(0.616)} & \multicolumn{1}{l}{(0.115)} & \multicolumn{1}{l}{(0.098)} & \multicolumn{1}{l}{(0.724)} & \multicolumn{1}{l}{(0.191)} \\ 
\hline
\multirow{2}{*}{PGD} & \textbf{0.688} & \textbf{0.748} & \textbf{11.20} & \textbf{0.205} & \textbf{2.459} & \textbf{3.391} & \textbf{0.430} & \textbf{0.231} & \textbf{1.798} & \textbf{0.555} \\
 & \multicolumn{1}{l}{\textbf{\textbf{(0.333)}}} & \multicolumn{1}{l}{\textbf{\textbf{(0.382)}}} & \multicolumn{1}{l}{\textbf{\textbf{(5.216)}}} & \multicolumn{1}{l}{\textbf{\textbf{(0.135)}}} & \multicolumn{1}{l}{\textbf{\textbf{(1.301)}}} & \multicolumn{1}{l}{\textbf{\textbf{(1.630)}}} & \multicolumn{1}{l}{\textbf{\textbf{(0.206)}}} & \multicolumn{1}{l}{\textbf{\textbf{(0.139)}}} & \multicolumn{1}{l}{\textbf{\textbf{(1.105)}}} & \multicolumn{1}{l}{\textbf{\textbf{(0.249)}}} \\
\bottomrule
\end{tabular}
\end{table*}

\subsubsection{\textbf{CLMPPCA (Black-box)}} We generated the attack vector using a CNN$+$LSTM surrogate model and evaluated the CLMPPCA model in a Black-box scenario. As with the other two Black-box experiments (i.e., MSCRED and MTAD-GAT), we saw a similar trend. The CLMPPCA model's prediction error does increase consistently for all subsystems when the attack vector is constructed using the surrogate model, as shown in Table~\ref{tab:clmppca}; in round brackets.
We can deduce from the CLMPPCA Black-box results that all three types of models investigated in this work (i.e., Autoencoder, DNNs, GNNs) are relatively equally susceptible to transferable adversarial attacks via surrogate models.

\subsection{\textbf{Summary of Results}}
Our findings indicate that the majority of SOTA anomaly detectors prioritized performance over robustness. This could have dire consequences if such systems are deployed in real-world systems. CLMPPCA is one such example, which is currently being deployed at KARI. Please note that we have informed KARI of the vulnerability in CLMPPCA; additional information is available in our Ethics Statement (see Section~\ref{sec:ethics}). Additionally, leveraging a surrogate model to conduct a Black-box attack can have a severe effect on the performance of the victim model. However, there are several limitations to surrogate, which we shall discuss in Section~\ref{sec:limitations}.

%% file: sections/6_Discussion.tex
\section{Discussion}
\label{sec:discussion}

\textbf{Adversarial Time Series Defense. } 
Adversarial training is one of the most commonly used defense methods against adversarial examples. However, as ~\citet{adversarialTraining} suggest, training a network to withstand one type of attack may weaken it against others. Additionally, \citet{DefenseIsWeak} outline various methods to conduct an adaptive attack and demonstrate that none of the 13 recently developed defense methods can withstand all types of adaptive attacks. Recently, a few techniques for defending against adversarial time series have been proposed. For example, \citet{APAE} propose an Approximate Projection Autoencoder (APAE) resistant to IFGSM attacks. However, it only considers autoencoder-based anomaly detectors. Moreover, the performance of several SOTA baselines reported in the paper is significantly lower than that reported in their original paper using the same publicly available benchmark dataset. As a result, a thorough examination of the defense methods is required. 

In order to encourage the studies of adversarial robustness for time series anomaly detection models, we will discuss here some possible approaches that are primarily motivated by computer vision areas. From the perspective of adversarial generation, perturbations created by attackers have mainly relied on gradients of model predictions \textit{w.r.t} its invaded inputs. We can apply the input-output Jacobian regularization in order to agnostically silent the model's gradients regardless of its input as was shown in \cite{Kenneth2021, Hoffman2019}.
On the other hand, when we have multiple classes in the training dataset, we can focus on aligning distributions of adversarial samples to be resembling to clean ones in the latent space, namely adversarial training \cite{Zhang2019defense, Bai2021, Bouniot2021}.  In the one-class training manner, we expect our defense model to learn the intrinsic representative features from the training dataset and be more robust to adding noise in the test set.  Therefore, regularizing the embedding space to be more compact is an appealing approach that so far has not been investigated in time-series anomaly detection areas thoroughly. This objective can be achieved via sparing the latent space with principal component analysis as demonstrated in \cite{Lai2019Robust, Lo2021Adver}.

\noindent
\textbf{Limitations and Future Work. }
\label{sec:limitations}
There are some limitations to our work, and future work will try to solve them. For instance, we could not evaluate all of the recent anomaly detectors in our work due to the following reasons: (i) The most important reason is that the codes are not publicly available in many cases or the code is outdated, making it hard to compare (we discuss this in detail in reproducibility section). (ii) It is hard to reproduce the same results as demonstrated by the paper, mainly when the codes are not from the original authors but developed by the community. Therefore, future work should look for more methods. Moreover, we have only applied FGSM, PGD, and SL1D (see Appendix~\ref{app_sec:MSCRED}) attacks on the detectors. We do provide results from other attacks such as Carlini-Wagner L2 and MIM on the UCR dataset on our GitHub repository. Another future work will be to transfer these and new adversarial attacks to anomaly detectors. 

We assumed that the training data for the surrogate model is either publicly available or obtained by probing the simulation results at multiple intelligently chosen places in the design parameter space. However, such an assumption may not hold true in a closed loop system. As a result, future research should focus on developing a more comprehensive strategy for acquiring training data for surrogate models.
Finally, developing robust detectors should be considered in future studies. 

\noindent
\textbf{New Adversarial Attacks. }
We have not developed a novel type of adversarial attack in this study and have instead utilised some of the more prevalent adversarial methods for the following reasons: (i) We believe that if a simple attack can demonstrate a system's vulnerability, then developing a new more complex attack solely to increase the novelty of the paper is futile, as the primary objective of this paper is to expose anomaly detector's vulnerabilities, not to develop new adversarial attacks. (ii) At the time the baselines reviewed in this study were published, the FGSM and PGD attacks were already well-known; thus, establishing that those baselines are not resilient against FGSM and PGD adversarial attacks provides a fair comparison.

\noindent
\textbf{Attack on Intrusion Detection System. } Intrusion detection is frequently associated with anomaly detection and, more broadly, novelty detection systems. In contrast to the realm of anomaly detection, numerous attempts have been made to investigate adversarial attacks against intrusion detection systems~\cite{IntrusionICLR1,IntrusionICLR2,IntrusionICLR3}. As a proof of concept, we also deployed similar adversarial attacks (i.e., FGSM and PGD) to an intrusion detection system for Controller Area Networks and discovered that the attacks are just as effective against them. Owing to the fact that this experiment requires extensive background information, and due to a shortage of space, we provide further details on our GitHub\footnote{{\color{RoyalBlue}\url{https://github.com/shahroztariq/Adversarial-Attacks-on-Timeseries}}} and more context here~\cite{tariq2022evaluating}.

\noindent
\textbf{Ethics Statement. }
\label{sec:ethics}
Our study, in our opinion, raises only one significant ethical issue (i.e., presenting the vulnerabilities of a deployed system). Now, we will describe how we deal with it. To begin, we downloaded the CLMPPCA code from GitHub. Second, we contacted the authors of the CLMPPCA paper and requested the dataset. Following KARI's security clearance. We were able to obtain access to the dataset and some code associated with the driver, which was kept private on purpose. We contacted the authors and informed them of our findings after identifying the vulnerabilities in CLMPPCA. The authors replicated our findings on the deployed system using the same attacks. For the time being, the system is offline, and the authors of the CLMPPCA paper and other KARI developers are investigating possible defense methods. We believe that adhering to this entire procedure resolves any ethical concerns regarding this matter.

%% file: sections/7_Conclusion.tex
\vspace{-5pt}
\section{Conclusion}
\label{sec:conclusion}
The concept of adversarial attacks on deep learning models for time series anomaly detection was considered in this paper. We defined and adapted adversarial attacks initially proposed for image recognition to time series data. On several benchmark datasets, we demonstrated how adversarial perturbations could reduce the accuracy of state-of-the-art anomaly detectors. As data scientists and developers increasingly implement deep neural network-based solutions for time series related real-world critical decision-making systems (e.g., in aerospace industries), we shed light on several critical use cases where adversarial attacks could have severe and dangerous repercussions. Additionally, we demonstrate empirically that White- and Black-box attacks are both conceivable and can result in significant performance deterioration. Finally, we discuss several defense strategies and possible future directions for adversarially resilient anomaly detector development.

\begin{acks}
This work was partially supported by the Basic Science Research Program through National Research Foundation of Korea (NRF) grant funded by the Korean Ministry of Science and ICT (MSIT) under No. 2020R1C1C1006004 and Institute for Information \& communication Technology Planning \& evaluation (IITP) grants funded by the Korean MSIT: (No. 2022-0-01199,  Graduate School of Convergence Security at Sungkyunkwan University), (No. 2022-0-01045, Self-directed Multi-Modal Intelligence for solving unknown, open domain problems), (No. 2022-0-00688, AI Platform to Fully Adapt and Reflect Privacy-Policy Changes), (No. 2021-0-02068, Artificial Intelligence Innovation Hub), (No. 2019-0-00421, AI Graduate School Support Program at Sungkyunkwan University), and (No. 2021-0-02309, Object Detection Research under Low Quality Video Condition).
\end{acks}

%% file: arxiv_sections/8_Appendix.tex

\section{Reproducibility}
\label{sec:repro}
\textbf{Issues in Baselines. } 
According to our research, the majority of recent anomaly detection methods do not make their source code publicly available. Additionally, many methods whose source code was made publicly available by their authors (or implemented unofficially) were outdated. As a result, we were unable to run them directly on the most recent machines. For instance, in our experiment, we used an Nvidia RTX 3090 GPU. We discovered that, due to some issues with the CUDA version, we could not run an older version of TensorFlow optimally. As a result, the code either takes an eternity to execute or does not execute at all.

\noindent
\textbf{Our Solution. } We chose to port the baselines to the latest versions of TensorFlow and PyTorch, respectively, which were 2.5.0 for TensorFlow and 1.9.0 for PyTorch at the time of our experiments. We used the CleverHans Library's \citep{CleverHans} FGSM, PGD, BIM, Carlini-Wagner L2, SL1D, and MIM attacks, which were recently ported to TensorFlow2 and PyTorch in version 4.0.0. As a result, our workflows are compatible with the latest libraries. Additionally, after cleaning the code, we will include some tutorial attacks (similar to those included in the CleverHans library for image datasets) that can be used to assess future detectors to adversarial attacks.

\noindent
\textbf{Guidelines for Baseline. } Note that it is difficult to port or implement all of the most recent methods on our own. Therefore, we tried our best with the limited resources that we had to make the baselines compatible with the latest version of libraries. We will provide some guidelines for creating new baselines and evaluating them against adversarial attacks on our GitHub page. We will leave it up to the community to add additional methods in the future.


\noindent
\textbf{Links to Baselines and Datasets. } We will include the updated codes for each baseline in our repository as well. We obtained the code of the baselines from the following repositories:
\begin{itemize}
    \item MSCRED~\cite{MSCRED}: {\footnotesize\url{https://github.com/Zhang-Zhi-Jie/Pytorch-MSCRED}}
    \item MTAD-GAT~\cite{MTADGAT}: {\footnotesize\url{https://github.com/ML4ITS/mtad-gat-pytorch}}
    \item CLMPPCA~\cite{CLMPPCA}: {\footnotesize\url{https://github.com/shahroztariq/CL-MPPCA}}
    \item CAN-ADF~\cite{CANADF,Shah1}: {\footnotesize\url{https://github.com/shahroztariq/CAN-ADF}}
    \item CANTransfer~\cite{CanTransfer}: {\footnotesize\url{https://github.com/shahroztariq/CANTransfer}}
\end{itemize}
Note that we are unable to share the KARI dataset as it is proprietary and requires security clearance to access. The link to rest of the dataset used in our evaluation are as follows:
\begin{itemize}
    \item SMAP and MSL: {\footnotesize \url{https://s3-us-west-2.amazonaws.com/telemanom/data.zip}}
    \item SMD: {\footnotesize\url{https://github.com/ML4ITS/mtad-gat-pytorch/tree/main/datasets}}
    \item Synthetic: {\footnotesize\url{https://github.com/Zhang-Zhi-Jie/Pytorch-MSCRED}}
    \item OTIDS: {\footnotesize\url{https://ocslab.hksecurity.net/Dataset/CAN-intrusion-dataset}}
\end{itemize}


\begin{figure*}[t]
    \centering
    \begin{subfigure}[t]{0.5\linewidth}
        \centering
      \includegraphics[width=1\linewidth]{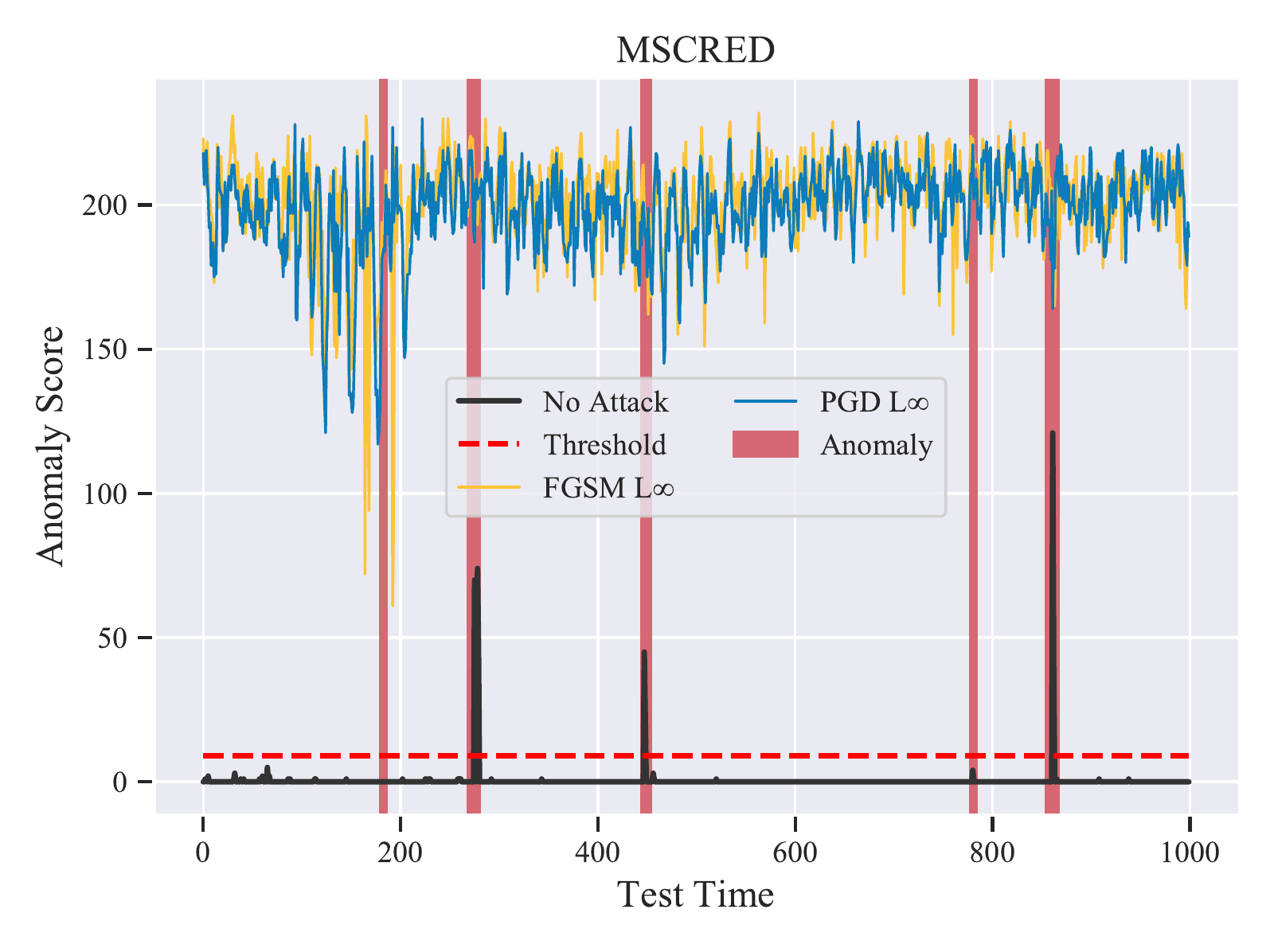}
    \caption{
    $\normmax$-norm based FGSM and PGD attacks.}
    \label{app_fig:MSCRED_Linf}
    \end{subfigure}\hfill
    \begin{subfigure}[t]{0.5\linewidth}
        \centering
    \includegraphics[width=1\linewidth]{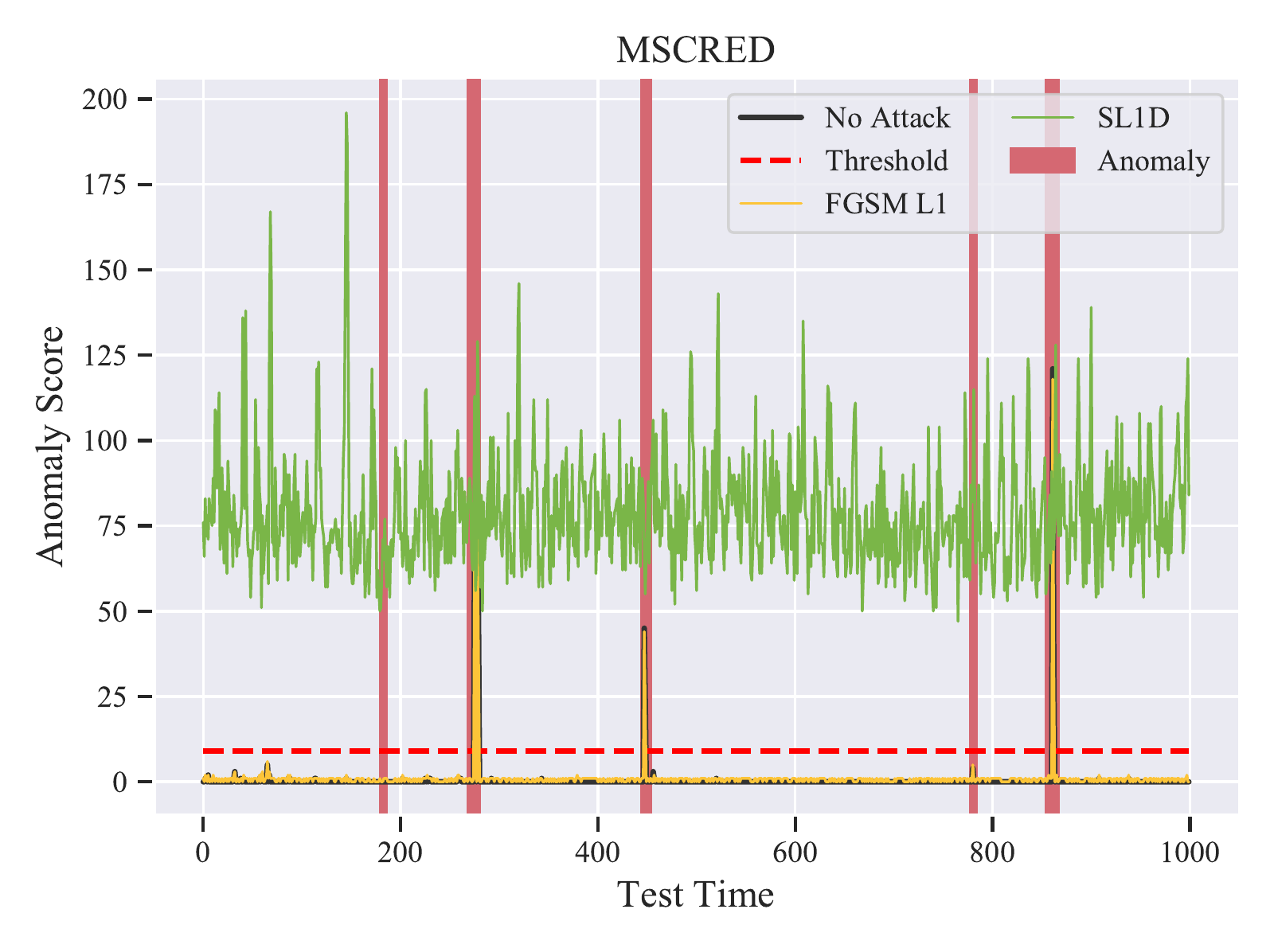}
    \caption{
    $\normlone$-norm based FGSM and SL1D attacks.}
    \label{app_fig:MSCRED_L1}
    \end{subfigure}\hfill
    \begin{subfigure}[t]{0.5\linewidth}
        \centering
    \includegraphics[width=1\linewidth]{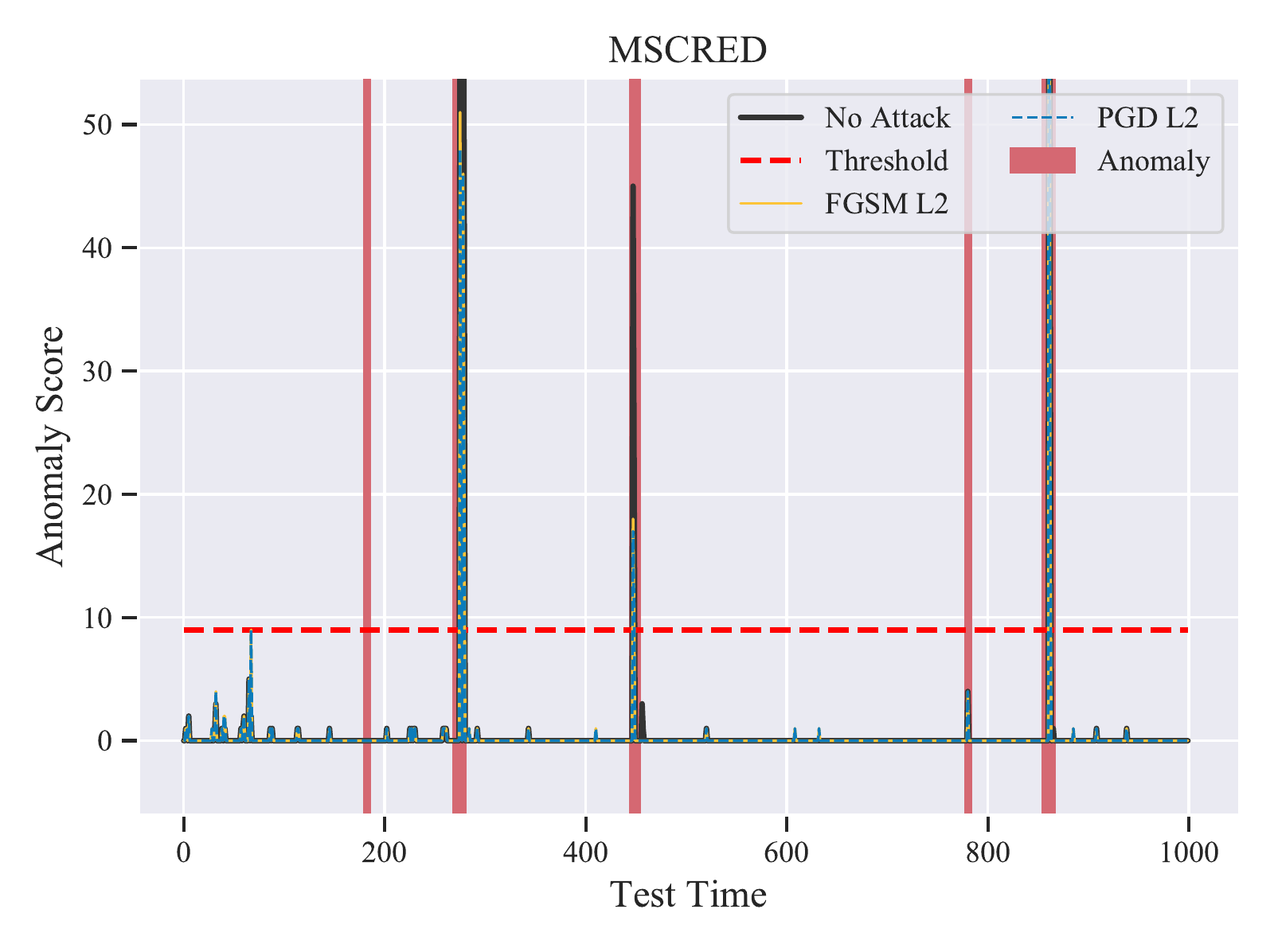}
    \caption{
    $\normltwo$-norm based FGSM and PGD attacks.}
    \label{app_fig:MSCRED_L2}
    \end{subfigure}
     \caption{Anomaly score comparison of MSCRED under No Attack, FGSM and PGD attacks.}
\end{figure*}

\begin{figure*}[t]
    \centering
    \begin{subfigure}[t]{1\linewidth}
    \includegraphics[trim={0pt 0pt 295pt 0pt},clip,width=1\linewidth]{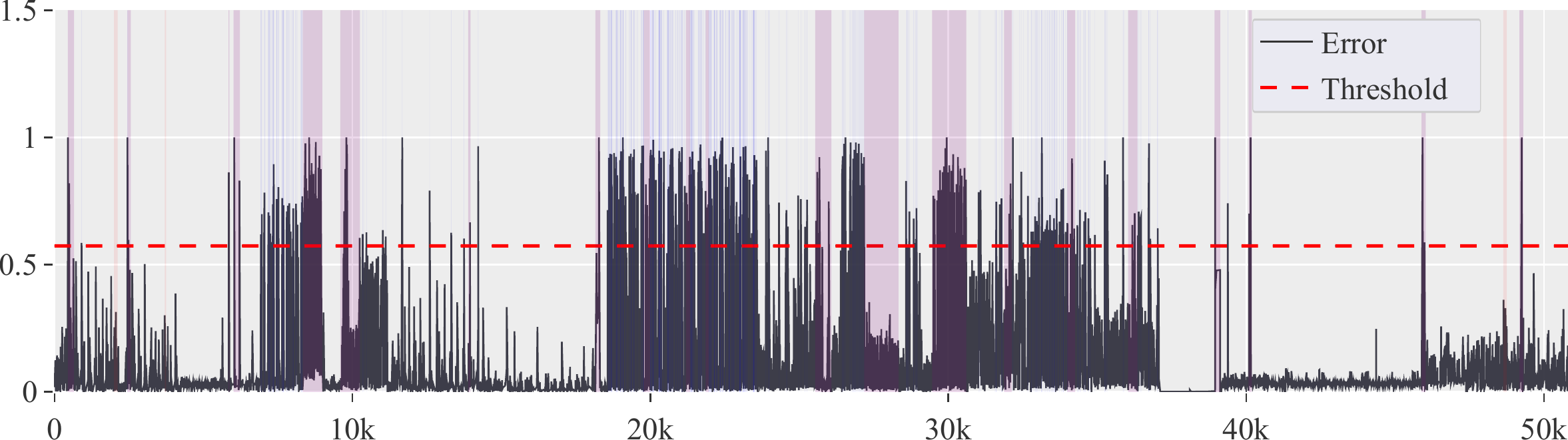}
    \caption{
    Normal data (i.e., No Attack).}
    \label{app_fig:MTAD-GAT_MSL_org}
    \end{subfigure}\hfill
    \begin{subfigure}[t]{1\linewidth}
        \centering
    \includegraphics[trim={0pt 0pt 295pt 0pt},clip,width=1\linewidth]{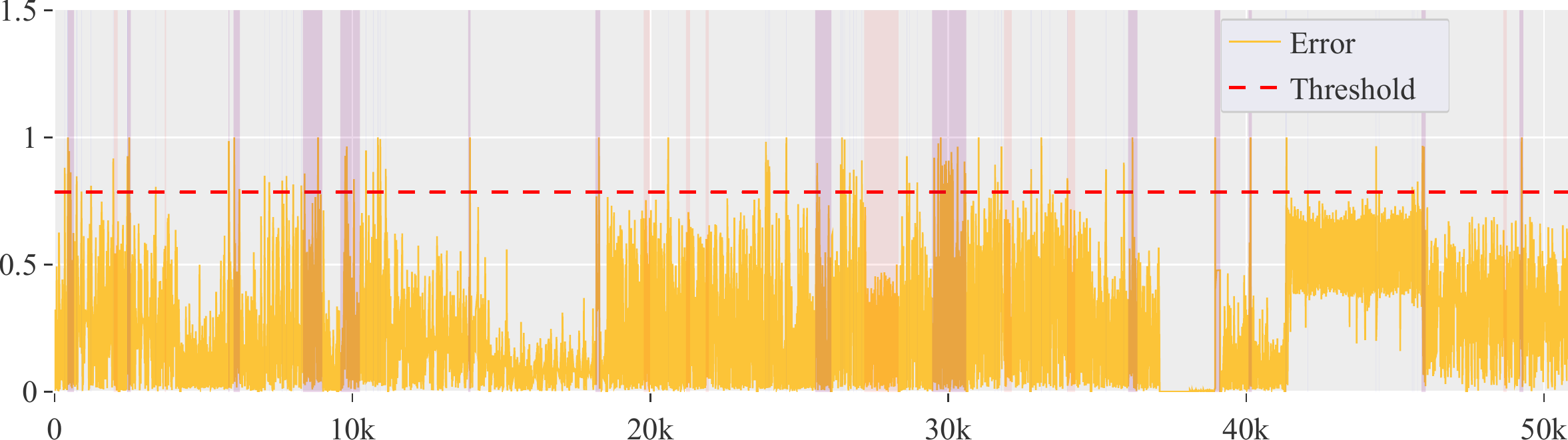}
    \caption{
    FGSM attack.}
    \label{app_fig:MTAD-GAT_MSL_fgsm}
    \end{subfigure}\hfill
    \begin{subfigure}[t]{1\linewidth}
        \centering
    \includegraphics[trim={0pt 0pt 295pt 0pt},clip,width=1\linewidth]{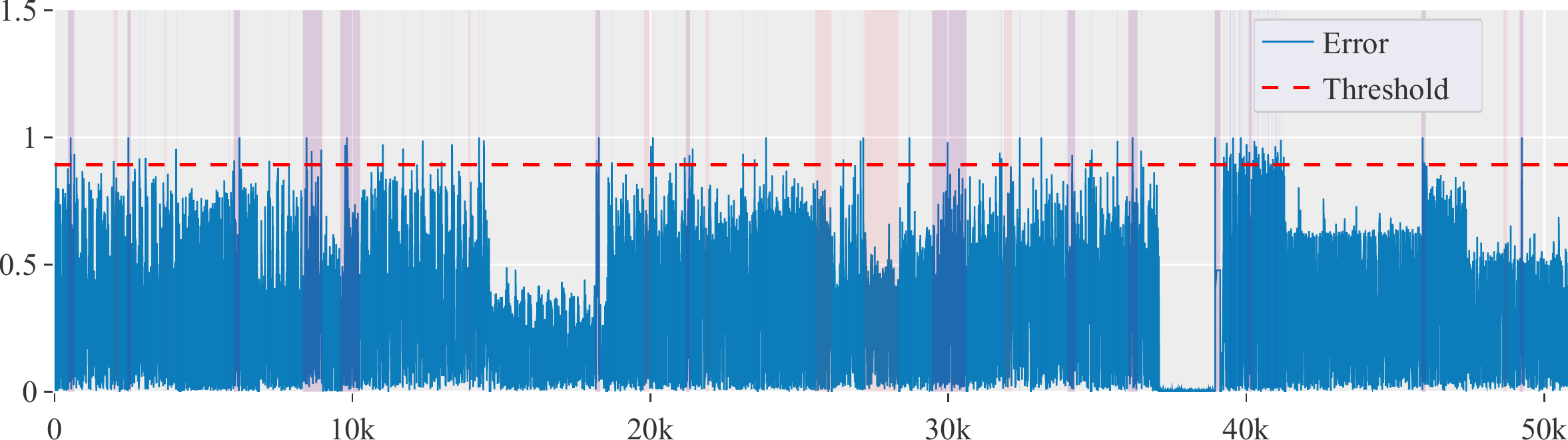}
    \caption{
    PGD attack.}
    \label{app_fig:MTAD-GAT_MSL_pgd}
    \end{subfigure}
    \caption{MTAD-GAT's anomaly score.}
\end{figure*}

\begin{figure*}[t]
    \begin{subfigure}[t]{0.5\linewidth}
        \centering
        \includegraphics[trim={650pt 0pt 650pt 60pt},clip,width=0.98\linewidth]{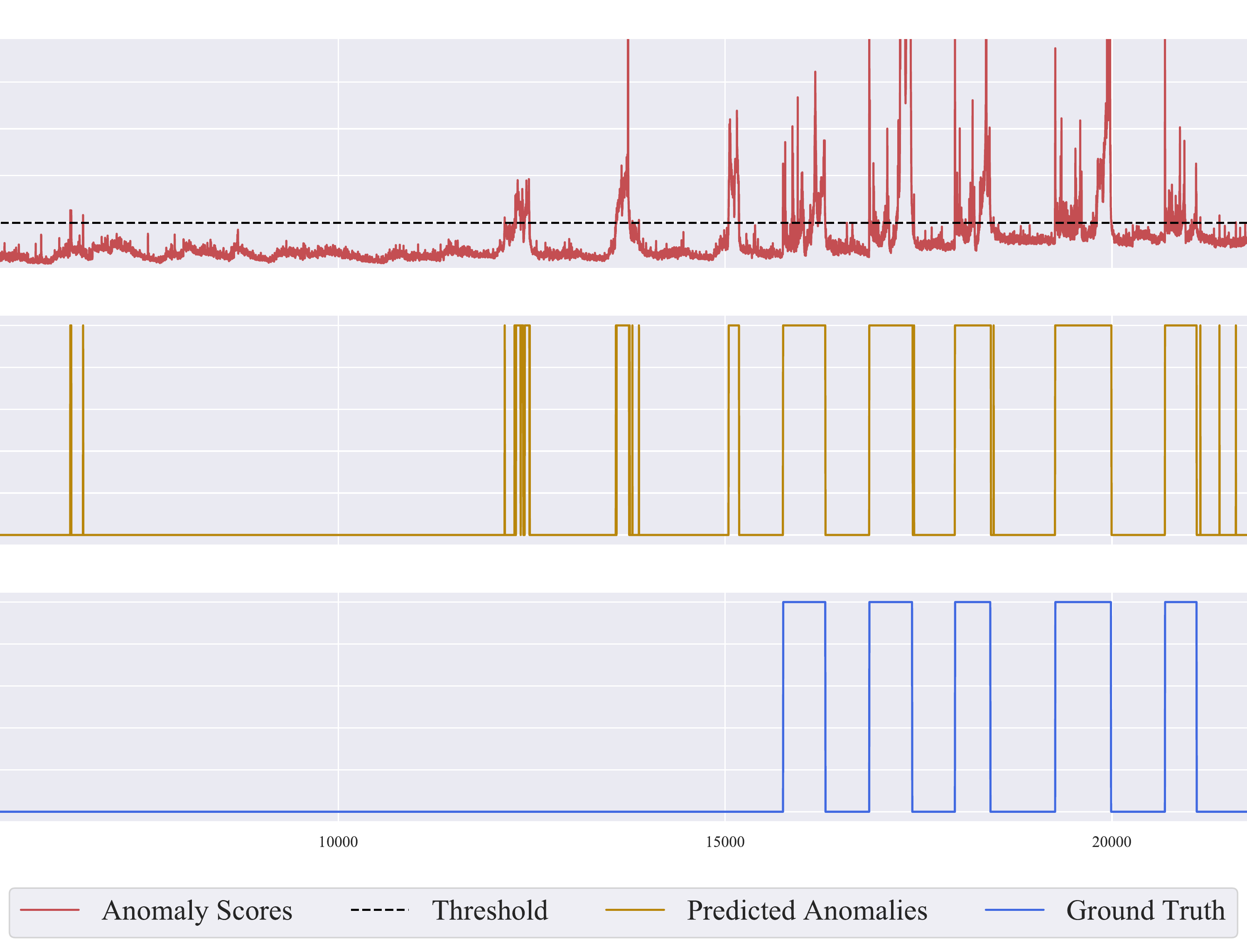}
        \caption{Normal data (i.e., No Attack).}
        
        \label{app_fig:MTAD_SMD_full_org}
    \end{subfigure}\hfill
    \begin{subfigure}[t]{0.5\linewidth}
                 \centering
        \includegraphics[trim={650pt 0pt 650pt 60pt},clip,width=0.98\linewidth]{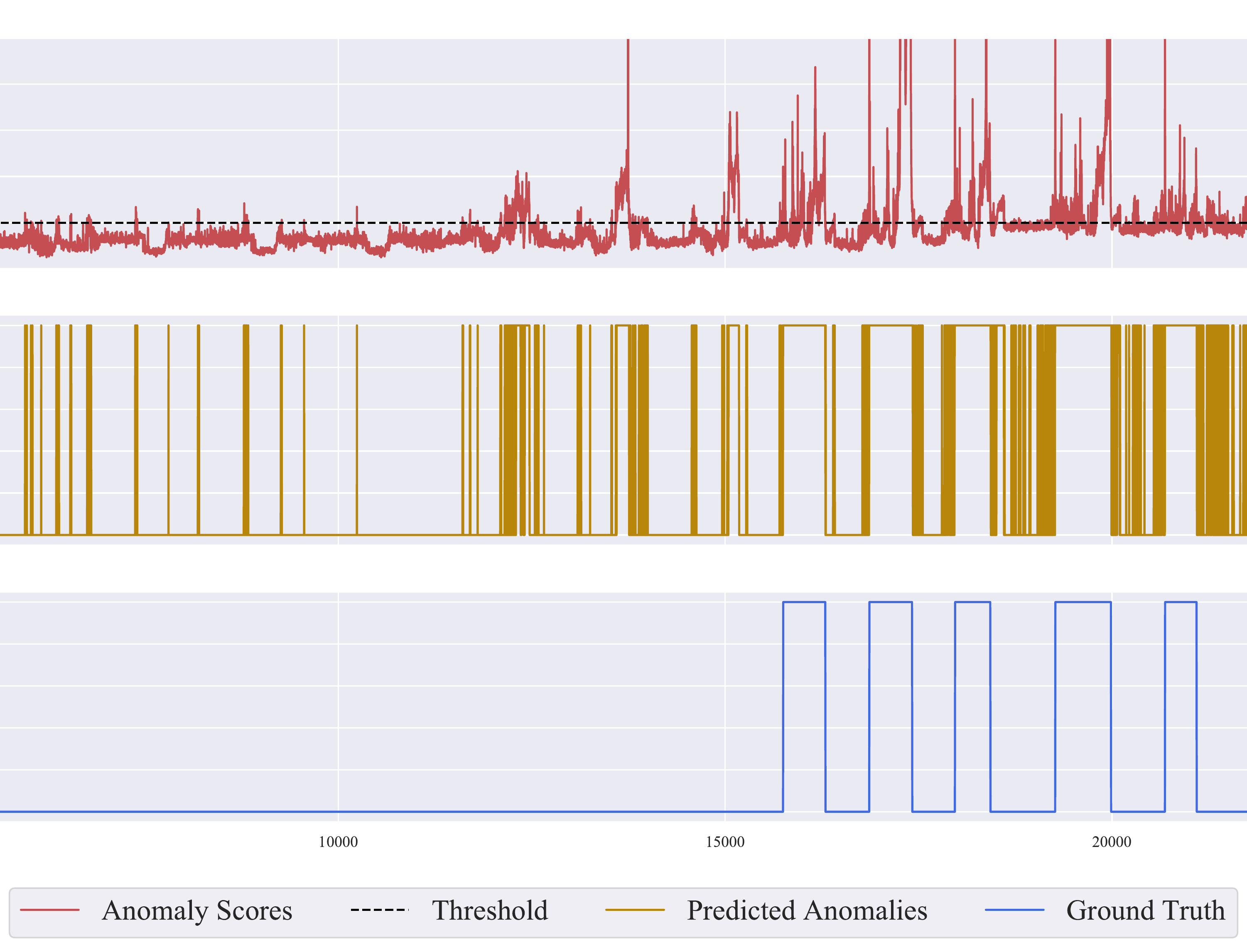}
        \caption{FGSM}
        \label{app_fig:MTAD_SMD_full_FGSM}
    \end{subfigure}\hfill
    \begin{subfigure}[t]{0.5\linewidth}
        \centering
        \includegraphics[trim={650pt 0pt 650pt 60pt},clip,width=0.98\linewidth]{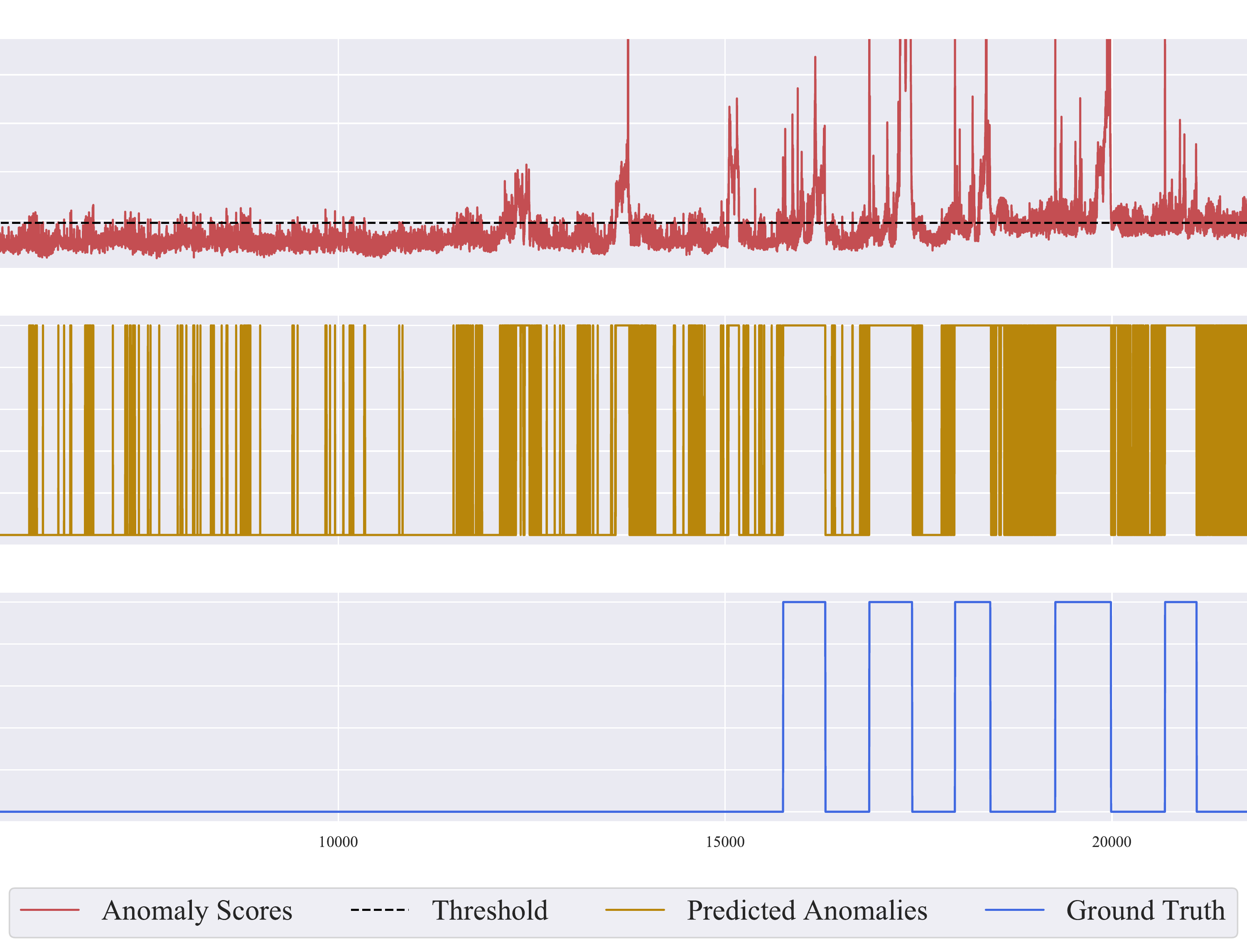}
        \caption{PGD}
        \label{app_fig:MTAD_SMD_full_PGD}
    \end{subfigure}
     \caption{The anomaly score and predicted anomalies for MTAD-GAT on SMD dataset.}
\end{figure*}

\begin{figure*}[t]
    \centering
    \begin{subfigure}[t]{1\linewidth}
        \includegraphics[trim={0pt 0pt 180pt 0pt},clip,width=0.98\linewidth]{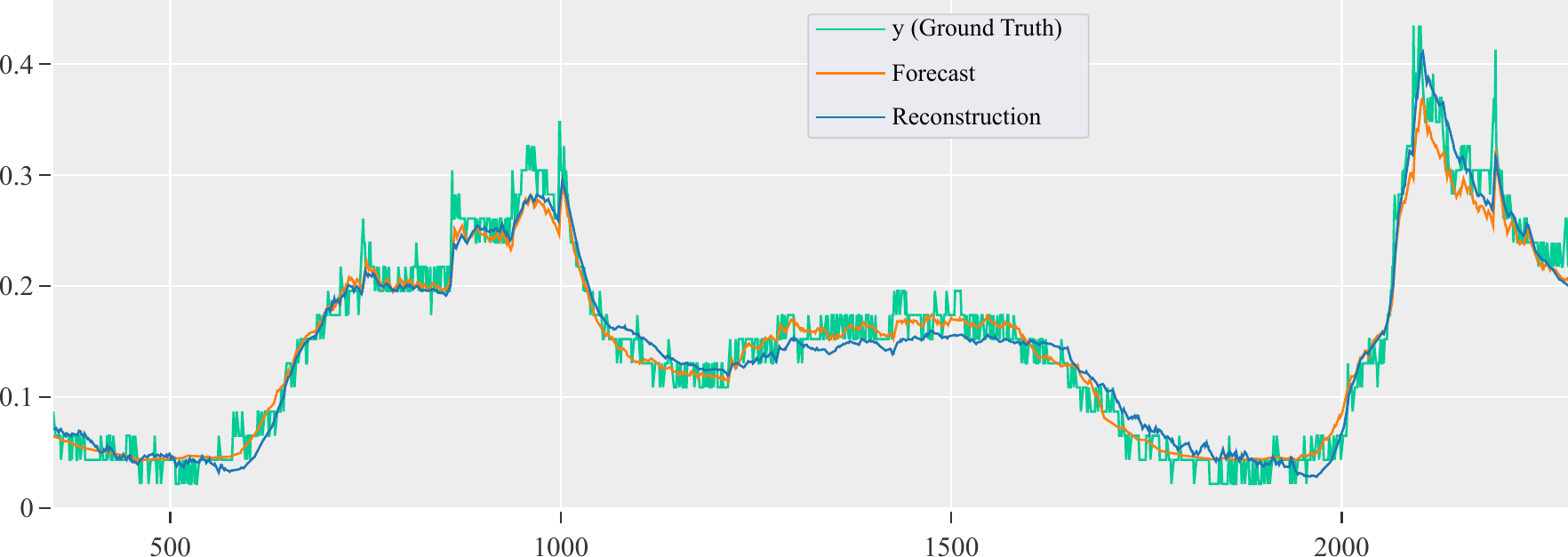}
    \caption{No Attack.}
    \label{app_fig:MTAD_SMD_Features_org}
    \end{subfigure}\hfill
    \begin{subfigure}[t]{1\linewidth}
        \centering
    \includegraphics[trim={0pt 0pt 180pt 0pt},clip,width=0.98\linewidth]{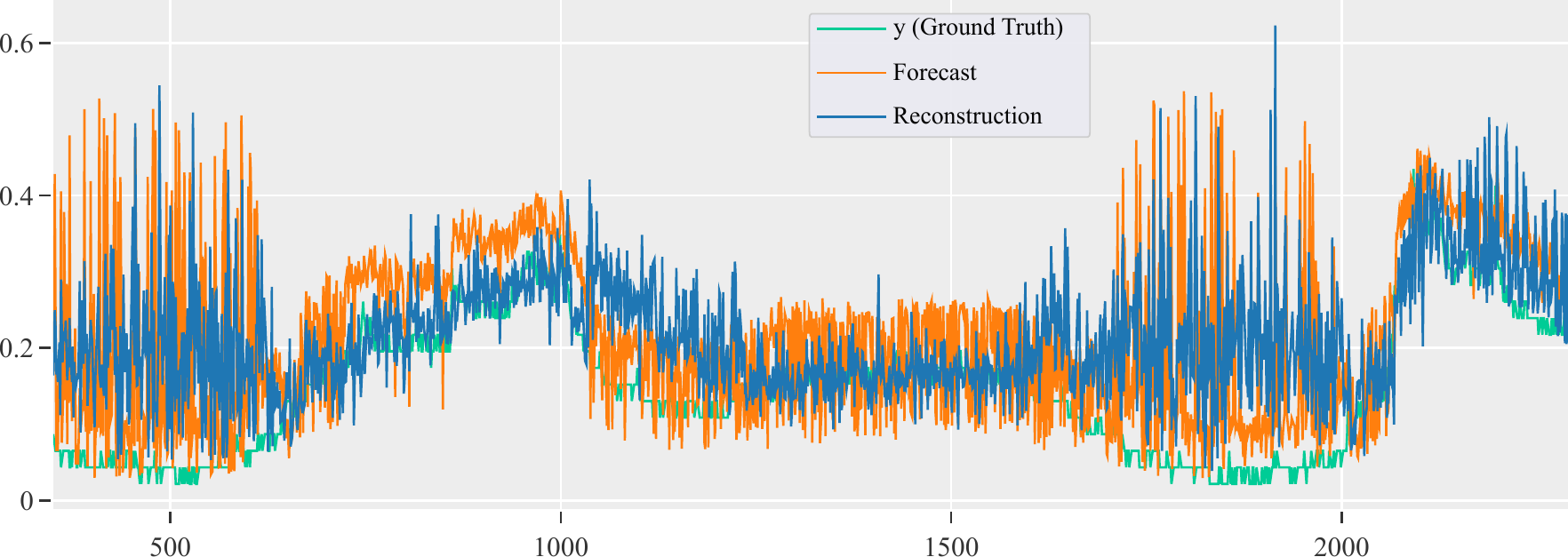}
    \caption{FGSM Attack.}
    \label{app_fig:MTAD_SMD_Features_fgsm}
    \end{subfigure}\hfill
    \begin{subfigure}[t]{1\linewidth}
    \centering
    \includegraphics[trim={0pt 0pt 180pt 0pt},clip,width=0.98\linewidth]{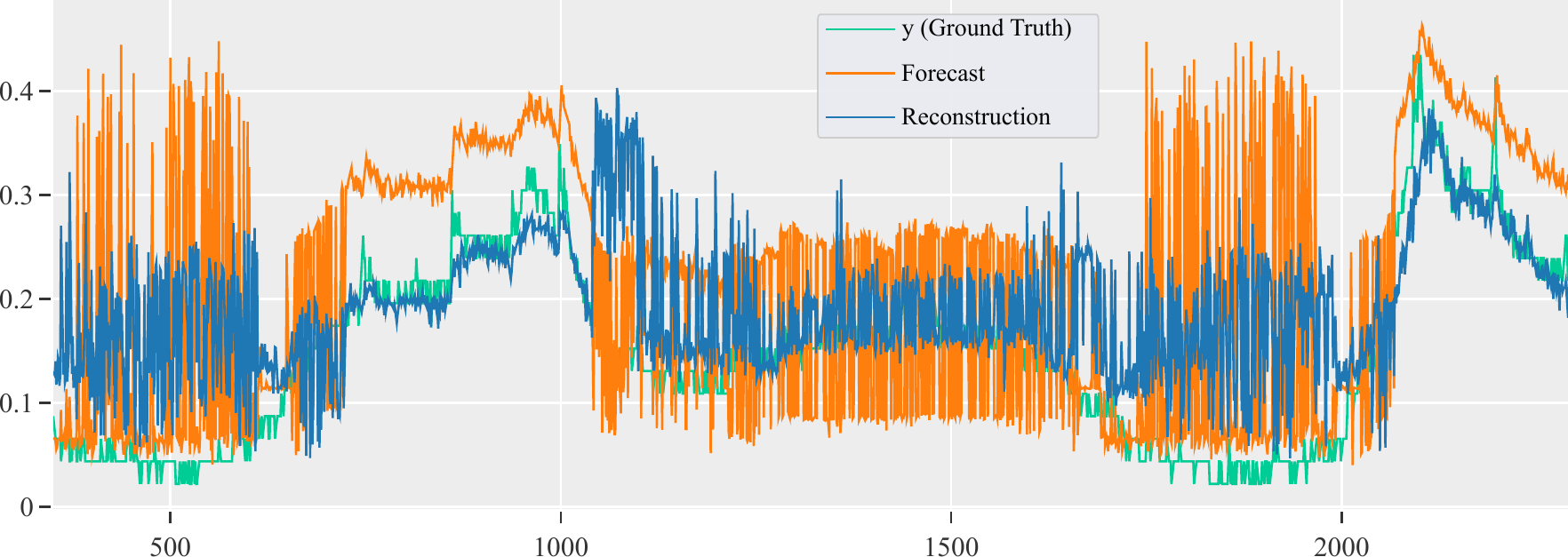}
    \caption{PGD Attack.}
    \label{app_fig:MTAD_SMD_Features_pgd}
    \end{subfigure}
    \caption{Comparison of Forecast and Reconstruction with $y_i$}
\end{figure*}

\begin{figure}[t]
\centering
    \begin{subfigure}[t]{1\linewidth}
        \centering
        \includegraphics[trim={0pt 0pt 0pt 0pt},clip,width=1\linewidth]{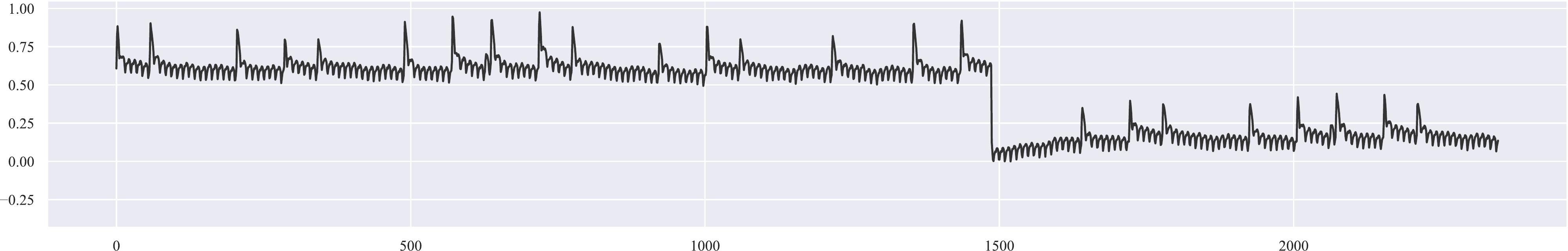}
        \caption{No Attack}
        \label{app_fig:org_vs_perturb_org}
    \end{subfigure}\hfill
    \begin{subfigure}[t]{1\linewidth}
       \centering
        \includegraphics[trim={0pt 0pt 0pt 0pt},clip,width=1\linewidth]{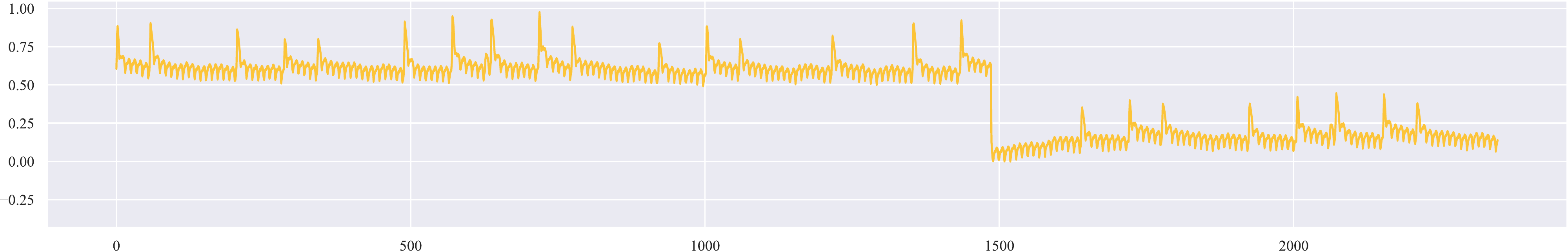}
        \caption{FGSM}
        \label{fig:my_labelapp_fig:org_vs_perturb_fgsm}
    \end{subfigure}\hfill
    \begin{subfigure}[t]{1\linewidth}
       \centering
        \includegraphics[trim={0pt 0pt 0pt 0pt},clip,width=1\linewidth]{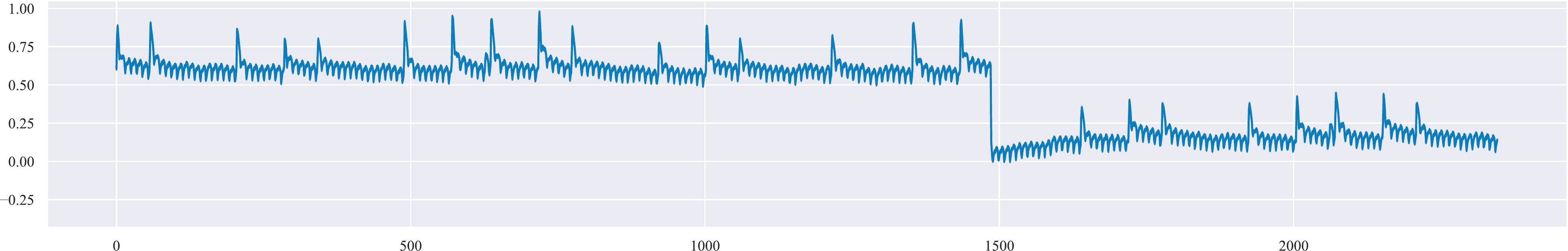}
        \caption{PGD}
        \label{app_fig:org_vs_perturb_pgd}
    \end{subfigure}\hfill
    \vspace{-10pt}
        \caption{A more detailed view of the same time series as in Figure~\ref{fig:clmppca}. }
        \label{app_fig:org_vs_perturb}
\end{figure}

\section{Supplementary Results: MSCRED}
\label{app_sec:MSCRED}
\noindent
\textbf{Details on $\normmax$ FGSM and PGD Attacks. }
In Figure~\ref{app_fig:MSCRED_Linf}, we detail MSCRED's performance against $\normmax$-norm FGSM and PGD attacks. Under normal conditions, we can see that the model correctly predicted three large anomalies but missed two minor ones. As a result, an F$_1$ score of $0.890$ is obtained. However, when attacked with either FGSM or PGD, the MSCRED model produces no meaningful results because it predicts everything as an anomaly. Furthermore, the patterns of anomaly score under FGSM and PGD attack are very similar to those observed during non-anomalous (or normal) periods. As a result, adjusting the threshold to account for changes in the anomaly score will not be as effective.


\noindent
\textbf{SL1D and FGSM $\normlone$ Attack. }
In Figure~\ref{app_fig:MSCRED_L1}, we present the results from two $\normlone$ attacks: (i) FGSM $\normlone$ and (ii) Sparse $\normlone$ Descent (SL1D) attacks. As discussed previously in the main paper, optimizing $\normlone$ and $\normltwo$-based attacks can be challenging. We can see an excellent illustration of this with the FGSM $\normlone$ attack, where adversarial examples from the $\normlone$-based FGSM attack produce nearly identical results to the No Attack data samples (with a few minor differences). However, the SL1D attack, also an $\normlone$-based attack, performs similar to the $\normmax$ attack discussed previously. Although the range of anomaly scores produced by SL1D attacks is slightly less than that produced by $\normmax$ attacks, it is still significantly higher than the threshold making the MSCRED model to predict the whole input time series as an anomaly.

\noindent
\textbf{$\normltwo$ FGSM and PGD Attack. }
The results of the $\normltwo$-based FGSM and PGD attacks are shown in Figure~\ref{app_fig:MSCRED_L2}. Almost identical to the $\normlone$-based FGSM attack, the $\normltwo$-based FGSM attack produces adversarial samples that have no effect on the anomaly score and are thus deemed ineffective. Similar results are obtained using the $\normltwo$-based PGD attack. As illustrated in Figure~\ref{app_fig:MSCRED_L2}, the Anomaly scores for No Attack, FGSM $\normltwo$, and PGD $\normltwo$ all overlap significantly.

\section{Supplementary Results: MTAD-GAT}
\label{app_sec:MTAD-GAT}
\noindent
\textbf{Detailed view of MTAD-GAT Results on MSL Dataset. }
In this section, we discuss the MTAD-GAT results on the MSL dataset in greater detail. Figure~\ref{app_fig:MTAD-GAT_MSL_org}--~\ref{app_fig:MTAD-GAT_MSL_pgd} show No Attack, FGSM attack, and PGD attack results on the entire test data, respectively. We can see that MTAD-GAT predicts fewer anomalies under FGSM and PGD attacks than normal conditions (i.e., No Attack), resulting in a higher rate of false negatives. We have now discussed both of these scenarios in detail in this work: (i) adversarial attack to generate false positives and (ii) adversarial attack to generate false negatives. Additionally, consistent with our previous findings, PGD performs better than FGSM and generates more false negatives than FGSM.




\noindent
\textbf{Results on SMD Dataset for MTAD-GAT. }
We present additional details on the MTAD-GAT results using the Server Machine Dataset (SMD) in Figure~\ref{app_fig:MTAD_SMD_full_org}, \ref{app_fig:MTAD_SMD_full_FGSM} and \ref{app_fig:MTAD_SMD_full_PGD} . In the figures, the top row (in red) represents the Anomaly scores, the middle row (in brown) represents the MTAD-GAT predictions, and the bottom row (in blue) represents the ground truth. We can see that MTAD-GAT performs at a state-of-the-art level under normal conditions. However, when subjected to FGSM and PGD attacks, it generates a large number of false positives, resulting in a significant decrease in overall performance. Also, we can observe that when PGD is used, MTAD-GAT produces more false positives than when FGSM is used.


\noindent
\textbf{Effects of FGSM and PGD attacks on MTAD-GAT's Features. }
As previously stated, MTAD-GAT is composed of two components (i.e., forecasting and reconstruction). We demonstrate in Figure~\ref{app_fig:MTAD_SMD_Features_org}--~\ref{app_fig:MTAD_SMD_Features_pgd} that both components become equally ineffective when subjected to adversarial attacks. For example, in normal circumstances (as illustrated in Figure~\ref{app_fig:MTAD_SMD_Features_org}), the forecast and reconstruction are quite close to the $y_i$ (ground truth). However, when attacked by FGSM, they deviate from the ground truth, fooling the system into believing it is an anomaly. Additionally, forecast and reconstruction are more chaotic during a PGD attack. As a result, detection performance is even lower than that of a FGSM attack.

    

    

\section{Supplementary Results: CLMPPCA}
\label{app_sec:CLMPPCA}
\noindent
\textbf{Original vs. Perturbed Samples. }
We compare some samples of original and perturbed time series in this section. The ground truth (in black), the FGSM (in yellow), and the PGD (in Blue) are depicted in Figure~\ref{app_fig:org_vs_perturb}. We can easily see that all three of the time series overlap, rendering them largely indistinguishable to the naked eye. Additionally, Figure~\ref{app_fig:org_vs_perturb_org}--~\ref{app_fig:org_vs_perturb_pgd} show an expanded version of the time series depicted in Figure~\ref{fig:clmppca}. Each of the three time series (i.e., No Attack, FGSM, and PGD) appears identical. Here, we demonstrate that even simpler adversarial attacks such as FGSM and PGD can be highly effective on time series data. Such perturbations will go unnoticed by a human observer.


\section{UCR Dataset Results}
\label{app_sec:UCR}
In addition to all the experiment on state-of-the-art anomaly and intrusion detection system. We also cover general time series classification task where we attack a multilayer perception (MLP), a fully convolutional network and ResNet trained on different dataset from the UCR repository. We conduct an analysis of 71 datasets from the University of California, Riverside (UCR) repository. In future work, we will expand on this experiment by including additional neural networks (MobileNet, EfficientNet, DenseNet, and Inception Time) and datasets (the remainder of the UCR dataset, datasets from the UEA repository). 

\begin{figure*}[t]
\centering
\centering
    \begin{subfigure}[t]{0.32\linewidth}
        \centering
        \includegraphics[trim={25pt 0pt 25pt 0pt},clip,width=1\linewidth]{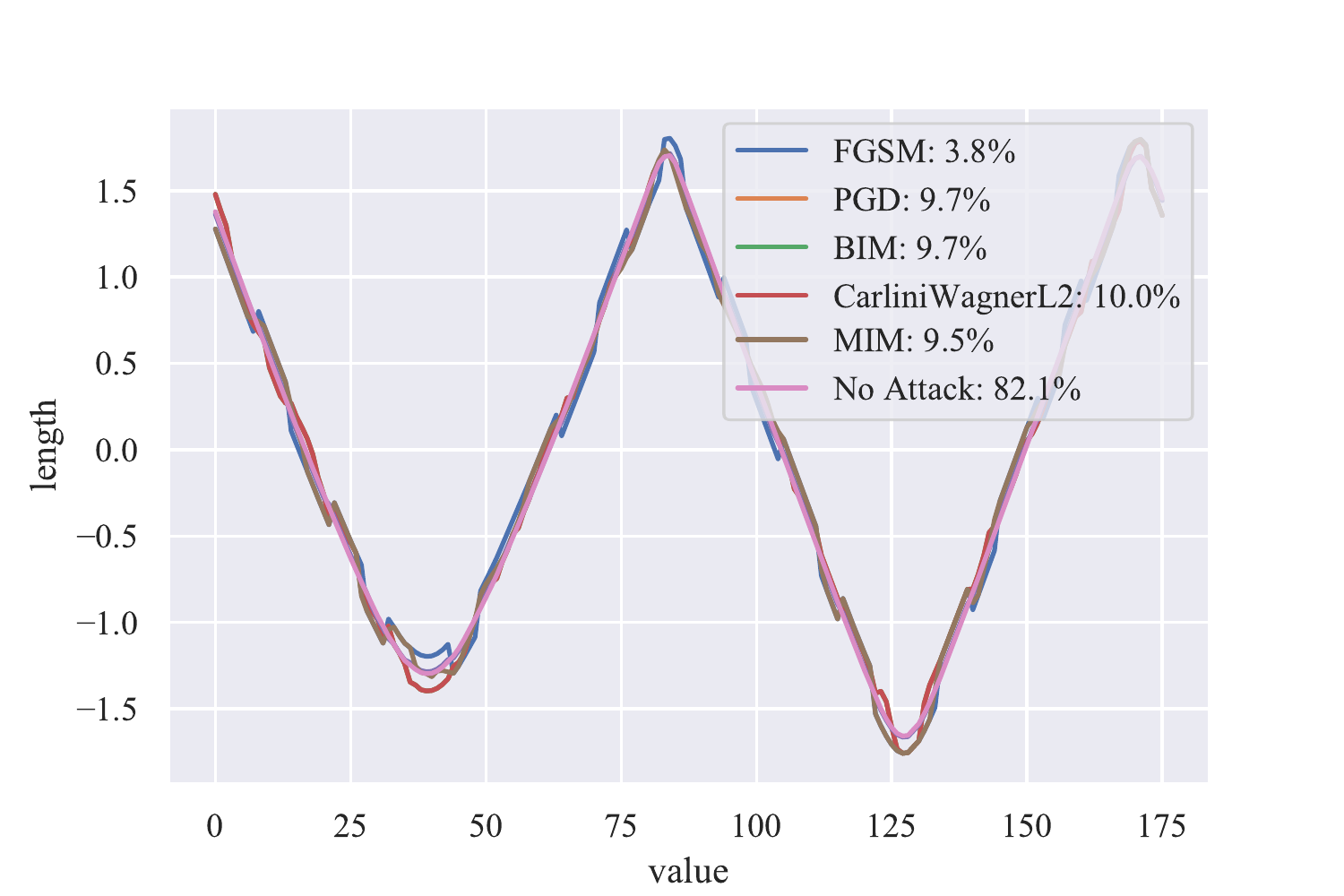}
        \caption{Adiac dataset}
    \end{subfigure}
    \begin{subfigure}[t]{0.32\linewidth}
        \centering
        \includegraphics[trim={25pt 0pt 25pt 0pt},clip,width=1\linewidth]{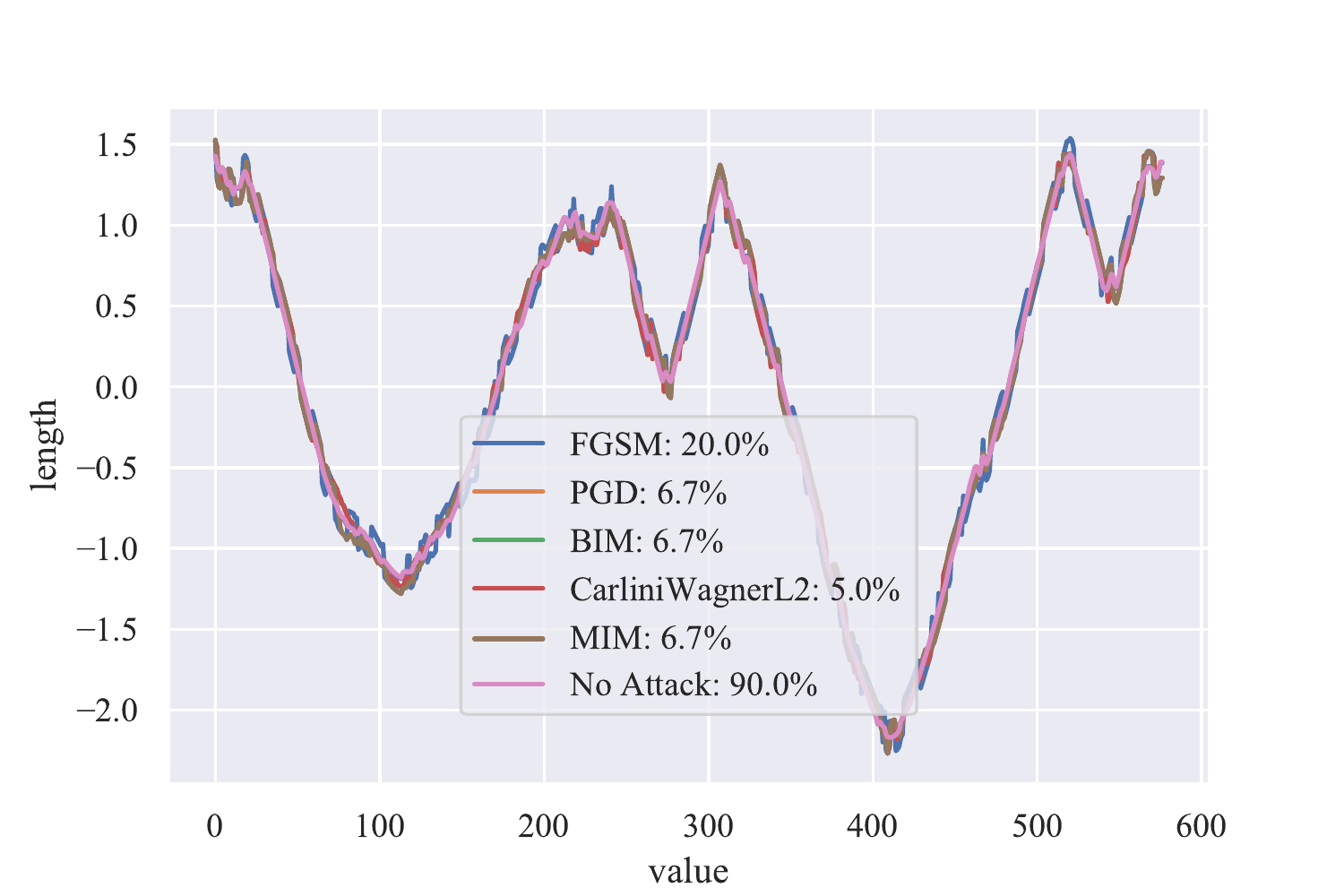}
        \caption{Car dataset}
    \end{subfigure}\hfill
    \begin{subfigure}[t]{0.32\linewidth}
        \centering
        \includegraphics[trim={25pt 0pt 25pt 0pt},clip,width=1\linewidth]{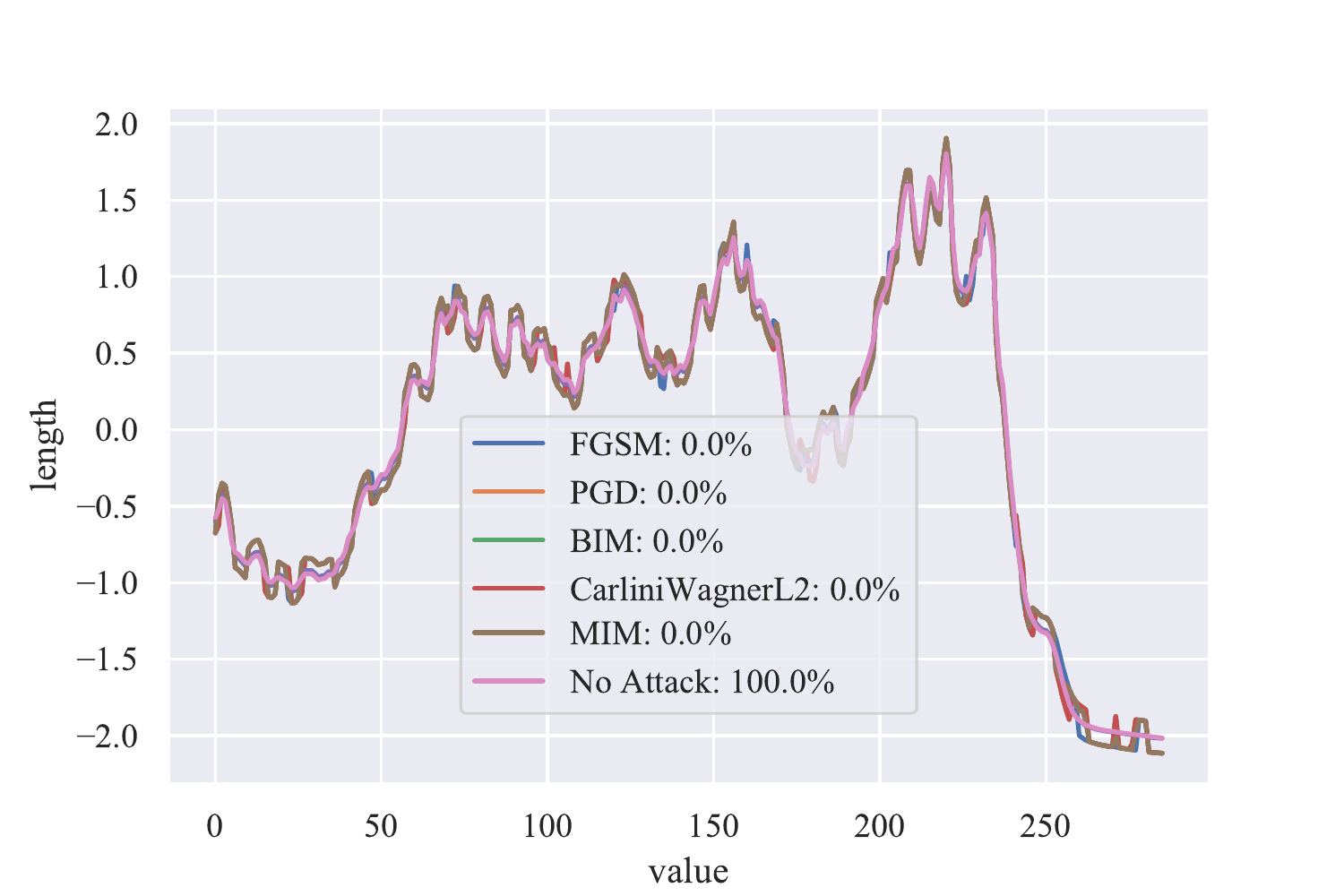}
        \caption{Coffee dataset}
    \end{subfigure}
    \begin{subfigure}[t]{0.32\linewidth}
        \centering
        \includegraphics[trim={25pt 0pt 25pt 0pt},clip,width=1\linewidth]{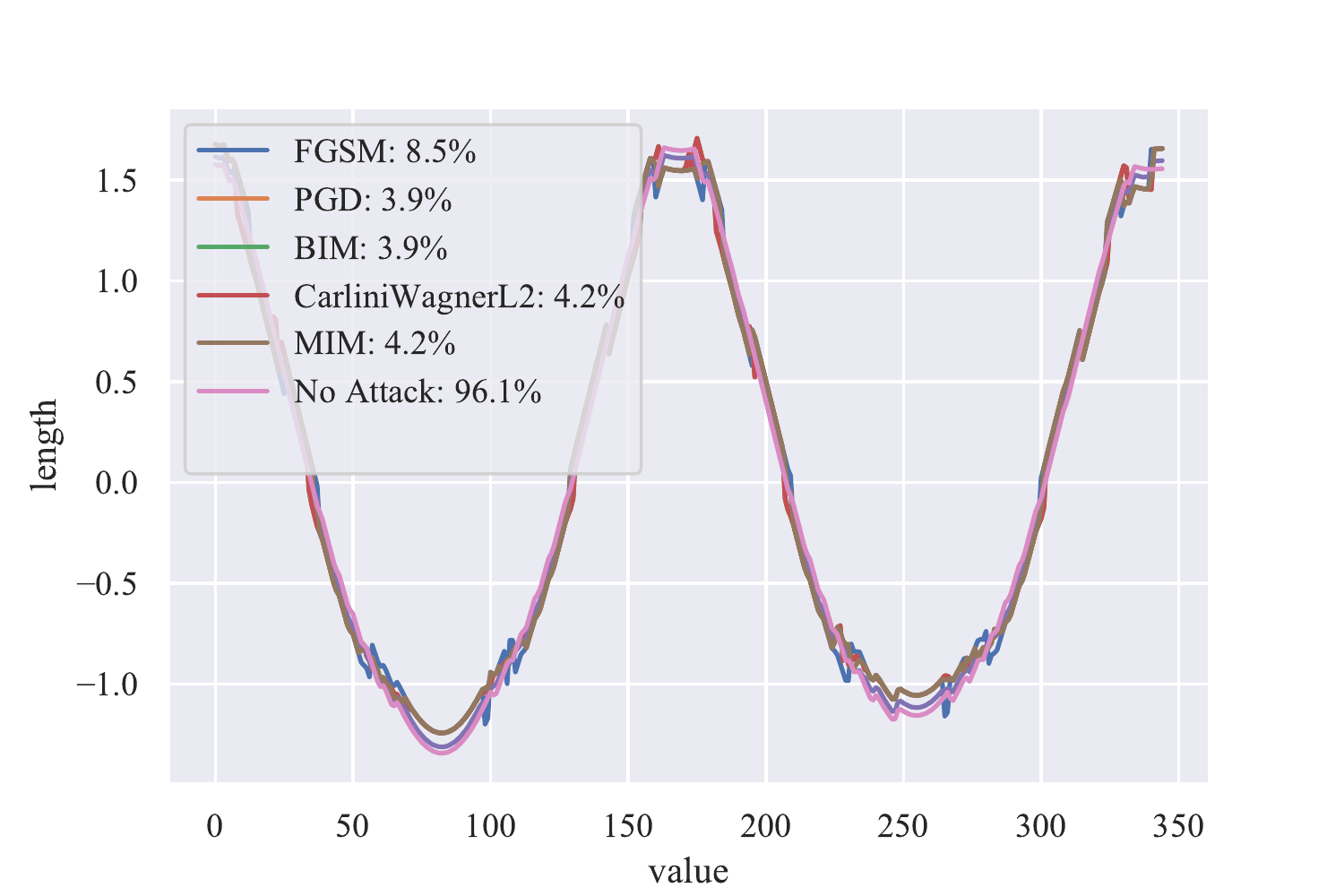}
        \caption{Diatom Size Reduction dataset}
    \end{subfigure}\hfill
    \begin{subfigure}[t]{0.32\linewidth}
        \centering
        \includegraphics[trim={25pt 0pt 25pt 0pt},clip,width=1\linewidth]{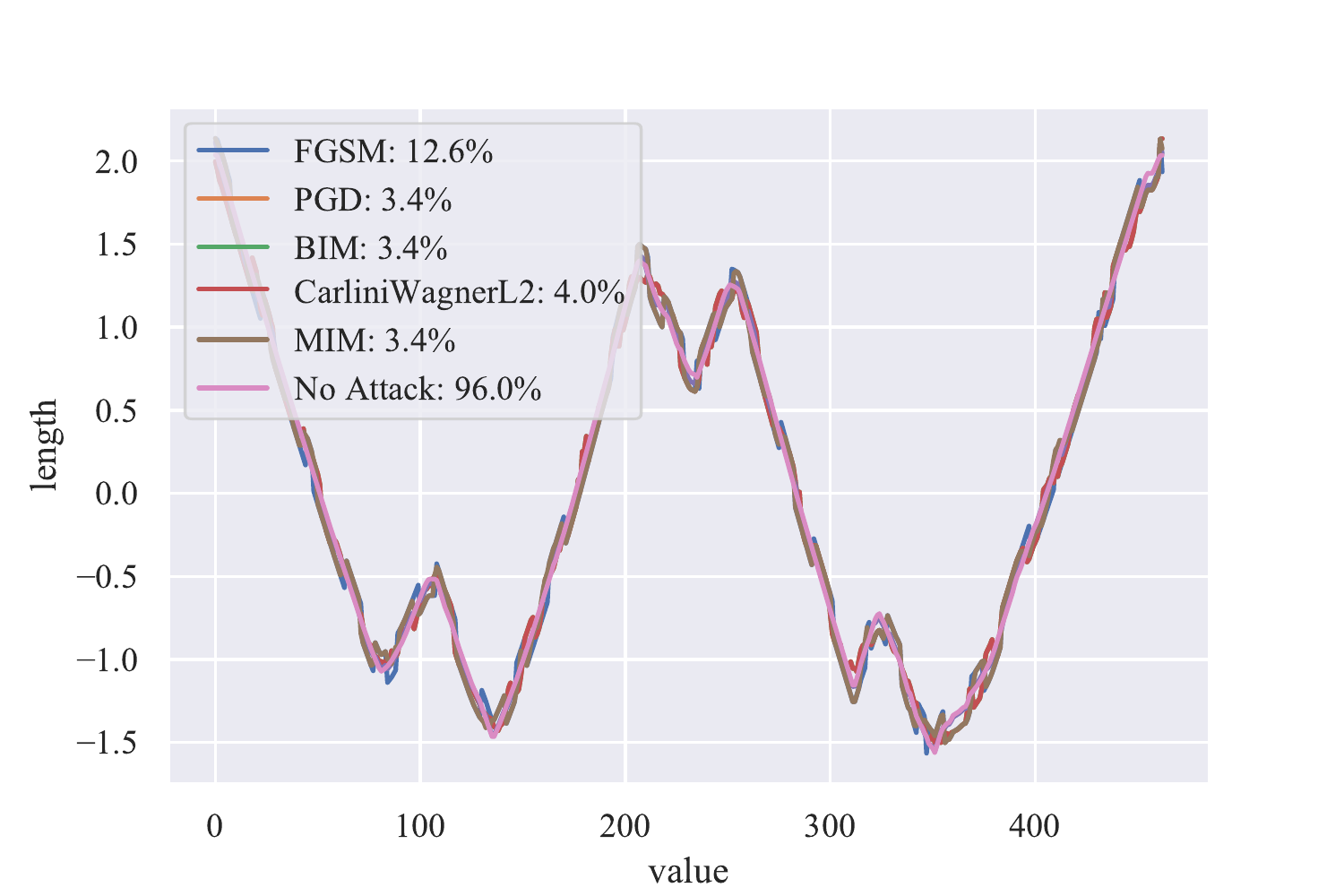}
        \caption{Fish dataset}
    \end{subfigure}
    \begin{subfigure}[t]{0.32\linewidth}
        \centering
        \includegraphics[trim={29pt 0pt 25pt 0pt},clip,width=1\linewidth]{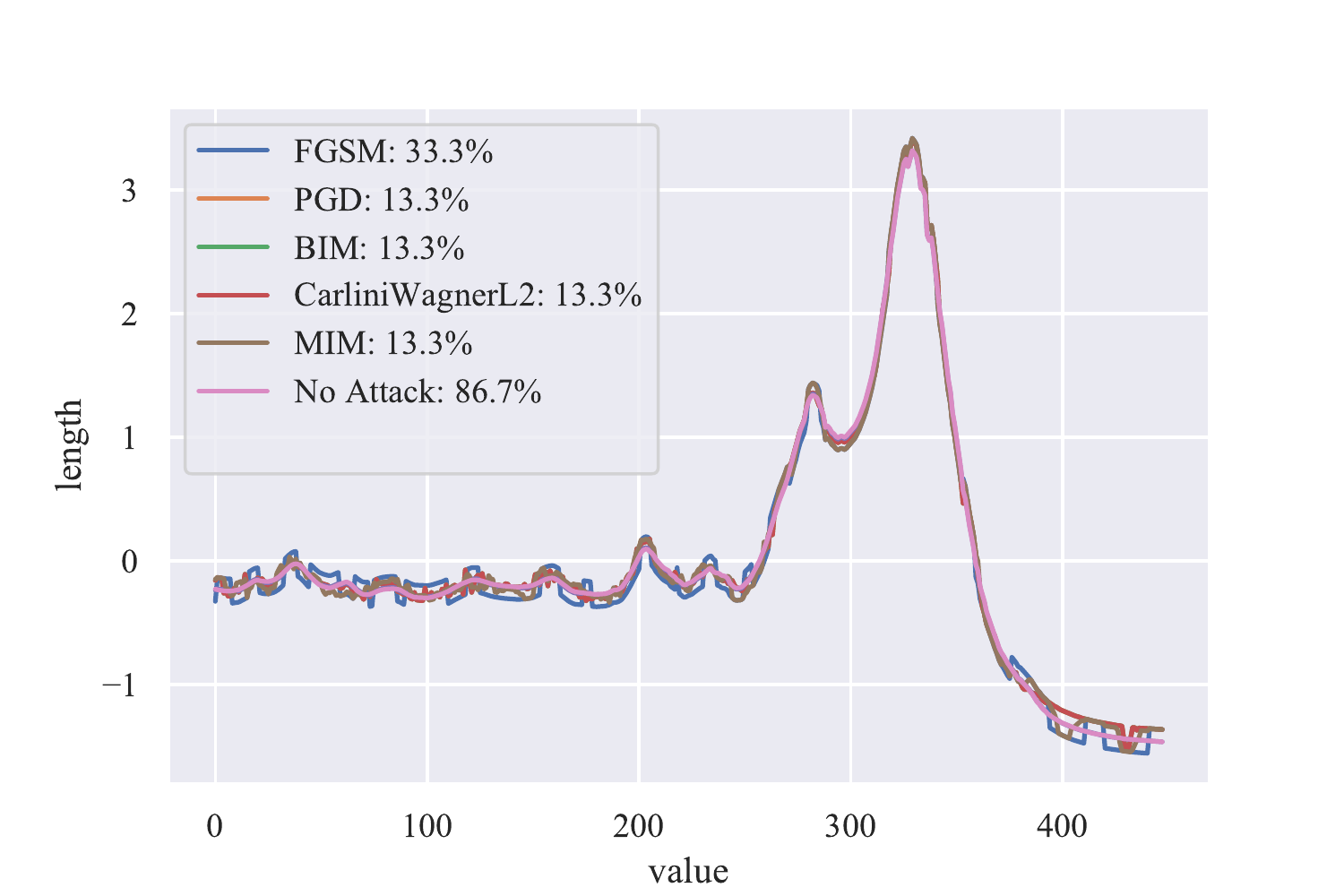}
        \caption{Meat dataset}
    \end{subfigure}\hfill
        \caption{Comparison of original vs. perturbed samples from different datasets inside UCR repository. Moreover, the classification results of ResNet under normal and five  different attack scenarios is also provided.}
        \label{app_fig:UCR}
\end{figure*}

We find that the Carlini-Wagner $\normltwo$ attack provides the best adversarial examples, resulting in the most significant performance degradation. In Figure~\ref{app_fig:UCR}, we show some original samples and the corresponding perturbed samples generated by FGSM, PGD, BIM, Carlini-Wagner $\normltwo$, and MIM attacks on UCR datasets. Additionally, we present the ResNet classification results in Figure~\ref{app_fig:UCR}. Finally, in Table~\ref{app_tab:MLP}--~\ref{app_tab:ResNet}, we present the classification results for MLP, FCN, and ResNet.

\newpage
\arrayrulecolor[rgb]{0.8,0.8,0.8}
\begin{longtable}{|l|r|r|r|r|r|r|} 
\caption{Multilayer perceptron (MLP) classification result on UCR repository under five adversarial attacks.}\label{app_tab:MLP}\\ 
\hline
\multicolumn{7}{|c|}{{\cellcolor[rgb]{0.388,0.824,0.592}}\textbf{MLP}} \endfirsthead 
\hline
\textbf{Datasets} & \multicolumn{1}{c|}{\textbf{FGSM}} & \multicolumn{1}{c|}{\textbf{PGD}} & \multicolumn{1}{c|}{\textbf{BIM}} & \multicolumn{1}{c|}{\begin{tabular}[c]{@{}c@{}}\textbf{Carlini}\\\textbf{Wagner}\\\textbf{$\normltwo$}\end{tabular}} & \multicolumn{1}{c|}{\textbf{MIM}} & \multicolumn{1}{c|}{\begin{tabular}[c]{@{}c@{}}\textit{No}\\\textit{Attack}\end{tabular}} \\ 
\hline
\rowcolor[rgb]{0.906,0.976,0.937} 50words & 44±0.8 & 42±1.3 & 42±1.3 & 35±1 & 43±1 & \textit{63±1.1} \\ 
\hline
Adiac & 14±1.8 & 15±1.6 & 15±1.6 & 16±1.3 & 16±1.8 & \textit{53±2.7} \\ 
\hline
\rowcolor[rgb]{0.906,0.976,0.937} ArrowHead & 29±3.9 & 27±3.3 & 27±3.3 & 24±4.3 & 27±3.1 & \textit{74±2.6} \\ 
\hline
Beef & 32±5.1 & 26±3.9 & 26±3.9 & 29±3.9 & 27±3.4 & \textit{78±3.9} \\ 
\hline
\rowcolor[rgb]{0.906,0.976,0.937} BeetleFly & 74±7.7 & 74±7.7 & 74±7.7 & 70±5 & 74±7.7 & \textit{75±13.3} \\ 
\hline
BirdChicken & 62±5.8 & 62±5.8 & 62±5.8 & 57±10.5 & 62±5.8 & \textit{69±2.9} \\ 
\hline
\rowcolor[rgb]{0.906,0.976,0.937} Car & 49±1 & 35±2.9 & 35±2.9 & 52±1 & 45±1 & \textit{83±1} \\ 
\hline
CBF & 76±2.6 & 76±2.4 & 76±2.4 & 63±3.5 & 76±2.6 & \textit{94±2.6} \\ 
\hline
\rowcolor[rgb]{0.906,0.976,0.937} \begin{tabular}[c]{@{}>{\cellcolor[rgb]{0.906,0.976,0.937}}l@{}}Chlorine\\Concentration\end{tabular} & 24±0.3 & 24±0.5 & 24±0.5 & 24±0.7 & 24±0.4 & \textit{65±0.4} \\ 
\hline
Coffee & 9±2.1 & 9±2.1 & 9±2.1 & 9±4.2 & 9±2.1 & \textit{100±0} \\ 
\hline
\rowcolor[rgb]{0.906,0.976,0.937} Computers & 46±1.1 & 45±1.1 & 45±1.1 & 45±1.1 & 45±1.1 & \textit{58±0.9} \\ 
\hline
Cricket\_X & 26±0.7 & 25±0.7 & 25±0.7 & 21±1 & 26±0.9 & \textit{45±1} \\ 
\hline
\rowcolor[rgb]{0.906,0.976,0.937} Cricket\_Y & 30±0.8 & 29±1.7 & 29±1.7 & 24±0.6 & 29±1.7 & \textit{48±1.6} \\ 
\hline
Cricket\_Z & 32±0.7 & 30±1.1 & 30±1.1 & 25±1 & 31±0.2 & \textit{44±1.2} \\ 
\hline
\rowcolor[rgb]{0.906,0.976,0.937} \begin{tabular}[c]{@{}>{\cellcolor[rgb]{0.906,0.976,0.937}}l@{}}DiatomSize\\Reduction\end{tabular} & 40±1.2 & 37±1.4 & 37±1.4 & 31±4 & 38±1.5 & \textit{95±2.4} \\ 
\hline
\begin{tabular}[c]{@{}l@{}}DistalPhalanx\\OutlineAgeGroup\end{tabular} & 16±0.9 & 16±1 & 16±1 & 16±1 & 16±0.9 & \textit{83±0.8} \\ 
\hline
\rowcolor[rgb]{0.906,0.976,0.937} \begin{tabular}[c]{@{}>{\cellcolor[rgb]{0.906,0.976,0.937}}l@{}}DistalPhalanx\\OutlineCorrect\end{tabular} & 29±1.4 & 28±1.8 & 28±1.8 & 25±0.9 & 29±1.7 & \textit{77±0.9} \\ 
\hline
\begin{tabular}[c]{@{}l@{}}Distal\\PhalanxTW\end{tabular} & 13±0.8 & 12±1.1 & 12±1.1 & 12±0.9 & 12±1.2 & \textit{78±0.7} \\ 
\hline
\rowcolor[rgb]{0.906,0.976,0.937} Earthquakes & 69±1.5 & 69±1.5 & 69±1.5 & 52±4.2 & 69±1.5 & \textit{73±1.1} \\ 
\hline
ECG200 & 60±1.8 & 60±2.1 & 60±2.1 & 29±5.8 & 60±2.1 & \textit{84±0.6} \\ 
\hline
\rowcolor[rgb]{0.906,0.976,0.937} ECG5000 & 65±0.2 & 64±0.3 & 64±0.3 & 61±0.3 & 64±0.4 & \textit{93±0.2} \\ 
\hline
ECGFiveDays & 48±2.3 & 46±2.1 & 46±2.1 & 35±4.8 & 47±2.1 & \textit{95±3.3} \\ 
\hline
\rowcolor[rgb]{0.906,0.976,0.937} ElectricDevices & 22±0.4 & 21±0.5 & 21±0.5 & 21±0.6 & 21±0.6 & \textit{55±0.8} \\ 
\hline
FaceAll & 57±0.3 & 56±0.4 & 56±0.4 & 39±0.9 & 56±0.2 & \textit{74±0.6} \\ 
\hline
\rowcolor[rgb]{0.906,0.976,0.937} FaceFour & 79±2.4 & 77±2 & 77±2 & 76±1.8 & 79±1.4 & \textit{88±0.7} \\ 
\hline
FacesUCR & 67±1.7 & 63±1.6 & 63±1.6 & 55±1.6 & 65±1.8 & \textit{83±1.2} \\ 
\hline
\rowcolor[rgb]{0.906,0.976,0.937} FISH & 16±2.1 & 8±1.2 & 8±1.2 & 14±1.9 & 12±1.2 & \textit{85±0.4} \\ 
\hline
Gun\_Point & 48±6.1 & 47±6.2 & 47±6.2 & 34±5.4 & 47±6.2 & \textit{92±1.4} \\ 
\hline
\rowcolor[rgb]{0.906,0.976,0.937} Ham & 34±2.4 & 34±2.6 & 34±2.6 & 48±3.5 & 34±2.6 & \textit{70±2} \\ 
\hline
Haptics & 21±0.9 & 21±0.8 & 21±0.8 & 21±1.2 & 20±0.7 & \textit{41±0.7} \\ 
\hline
\rowcolor[rgb]{0.906,0.976,0.937} Herring & 50±1.9 & 50±1.9 & 50±1.9 & 50±1.9 & 50±1.9 & \textit{51±1.9} \\ 
\hline
InlineSkate & 21±1.1 & 19±0.8 & 19±0.8 & 20±1.4 & 20±1.4 & \textit{34±0.7} \\ 
\hline
\rowcolor[rgb]{0.906,0.976,0.937} \begin{tabular}[c]{@{}>{\cellcolor[rgb]{0.906,0.976,0.937}}l@{}}InsectWing\\beatSound\end{tabular} & 37±0.7 & 30±0.3 & 30±0.3 & 42±0.3 & 34±0.4 & \textit{62±0.7} \\ 
\hline
\begin{tabular}[c]{@{}l@{}}ItalyPower\\Demand\end{tabular} & 82±0.8 & 82±0.9 & 82±0.9 & 11±1.4 & 82±0.9 & \textit{96±0.2} \\ 
\hline
\rowcolor[rgb]{0.906,0.976,0.937} \begin{tabular}[c]{@{}>{\cellcolor[rgb]{0.906,0.976,0.937}}l@{}}LargeKitchen\\Appliances\end{tabular} & 33±2.2 & 32±1.3 & 32±1.3 & 34±0.6 & 33±2.1 & \textit{51±0.5} \\ 
\hline
Lighting2 & 70±2.6 & 70±2.6 & 70±2.6 & 58±3.8 & 70±2.6 & \textit{65±3.5} \\ 
\hline
\rowcolor[rgb]{0.906,0.976,0.937} Lighting7 & 53±4.2 & 53±3.7 & 53±3.7 & 35±3.7 & 53±3.7 & \textit{64±2.4} \\ 
\hline
Meat & 26±1 & 26±1 & 26±1 & 25±1.7 & 26±1 & \textit{74±1} \\ 
\hline
\rowcolor[rgb]{0.906,0.976,0.937} MedicalImages & 39±1.9 & 36±2.2 & 36±2.2 & 26±0.5 & 37±2.2 & \textit{67±0.5} \\ 
\hline
\begin{tabular}[c]{@{}l@{}}MiddlePhalanx\\OutlineAgeGroup\end{tabular} & 32±10.7 & 26±4.8 & 26±4.8 & 20±0.8 & 27±5.7 & \textit{73±1.5} \\ 
\hline
\rowcolor[rgb]{0.906,0.976,0.937} \begin{tabular}[c]{@{}>{\cellcolor[rgb]{0.906,0.976,0.937}}l@{}}MiddlePhalanx\\OutlineCorrect\end{tabular} & 46±1.5 & 46±1.6 & 46±1.6 & 45±1.5 & 46±1.6 & \textit{56±1.5} \\ 
\hline
\begin{tabular}[c]{@{}l@{}}Middle\\PhalanxTW\end{tabular} & 18±2.9 & 18±2.8 & 18±2.8 & 18±1.7 & 18±2.9 & \textit{56±2.4} \\ 
\hline
\rowcolor[rgb]{0.906,0.976,0.937} MoteStrain & 79±0.7 & 79±0.7 & 79±0.7 & 53±2.3 & 79±0.7 & \textit{84±1.1} \\ 
\hline
OliveOil & 28±2 & 28±2 & 28±2 & 28±2 & 28±2 & \textit{59±2} \\ 
\hline
\rowcolor[rgb]{0.906,0.976,0.937} OSULeaf & 29±0.7 & 29±1.1 & 29±1.1 & 29±0.9 & 30±0.7 & \textit{45±0.3} \\ 
\hline
\begin{tabular}[c]{@{}l@{}}Phalanges\\OutlinesCorrect\end{tabular} & 33±3.2 & 33±2.6 & 33±2.6 & 33±2.3 & 33±2.7 & \textit{68±2.4} \\ 
\hline
\rowcolor[rgb]{0.906,0.976,0.937} Plane & 89±2 & 87±1.1 & 87±1.1 & 85±4.3 & 88±1.1 & \textit{98±1.1} \\ 
\hline
\begin{tabular}[c]{@{}l@{}}ProximalPhalanx\\OutlineAgeGroup\end{tabular} & 18±2 & 18±2.3 & 18±2.3 & 18±1.8 & 18±2.3 & \textit{81±1.9} \\ 
\hline
\rowcolor[rgb]{0.906,0.976,0.937} \begin{tabular}[c]{@{}>{\cellcolor[rgb]{0.906,0.976,0.937}}l@{}}ProximalPhalanx\\OutlineCorrect\end{tabular} & 36±1.4 & 34±1.1 & 34±1.1 & 33±1.6 & 34±0.9 & \textit{68±1.6} \\ 
\hline
\begin{tabular}[c]{@{}l@{}}Proximal\\PhalanxTW\end{tabular} & 41±3.9 & 42±4 & 42±4 & 42±4 & 42±3.9 & \textit{53±4.1} \\ 
\hline
\rowcolor[rgb]{0.906,0.976,0.937} \begin{tabular}[c]{@{}>{\cellcolor[rgb]{0.906,0.976,0.937}}l@{}}Refrigeration\\Devices\end{tabular} & 36±1.8 & 36±1.6 & 36±1.6 & 36±1.3 & 36±1.9 & \textit{43±1.2} \\ 
\hline
ScreenType & 39±1.4 & 38±1.8 & 38±1.8 & 38±1 & 39±1.6 & \textit{36±0.3} \\ 
\hline
\rowcolor[rgb]{0.906,0.976,0.937} ShapeletSim & 50±1.7 & 50±1.4 & 50±1.4 & 49±1.7 & 50±1.4 & \textit{48±0.9} \\ 
\hline
ShapesAll & 49±1.6 & 42±1.1 & 42±1.1 & 43±1.3 & 46±1.8 & \textit{70±0.2} \\ 
\hline
\rowcolor[rgb]{0.906,0.976,0.937} \begin{tabular}[c]{@{}>{\cellcolor[rgb]{0.906,0.976,0.937}}l@{}}SmallKitchen\\Appliances\end{tabular} & 33±1.4 & 34±1 & 34±1 & 36±1.6 & 34±1.1 & \textit{49±2.2} \\ 
\hline
\begin{tabular}[c]{@{}l@{}}SonyAIBO\\RobotSurface\end{tabular} & 68±2.6 & 68±2.6 & 68±2.6 & 62±7.3 & 68±2.6 & \textit{68±4.6} \\ 
\hline
\rowcolor[rgb]{0.906,0.976,0.937} \begin{tabular}[c]{@{}>{\cellcolor[rgb]{0.906,0.976,0.937}}l@{}}SonyAIBO\\RobotSurfaceII\end{tabular} & 81±0.8 & 81±0.8 & 81±0.8 & 71±0.6 & 81±0.8 & \textit{83±0.8} \\ 
\hline
Strawberry & 7±0.3 & 6±0.3 & 6±0.3 & 9±0.7 & 7±0.2 & \textit{96±0.3} \\ 
\hline
\rowcolor[rgb]{0.906,0.976,0.937} SwedishLeaf & 32±1.2 & 26±2.1 & 26±2.1 & 25±0.8 & 29±1.4 & \textit{82±0.3} \\ 
\hline
Symbols & 76±1.5 & 74±1.2 & 74±1.2 & 76±1.4 & 75±1 & \textit{89±0.2} \\ 
\hline
\rowcolor[rgb]{0.906,0.976,0.937} synthetic\_control & 80±1.6 & 80±1.7 & 80±1.7 & 37±3.6 & 80±1.6 & \textit{95±1} \\ 
\hline
ToeSegmentation1 & 51±1.5 & 51±1.5 & 51±1.5 & 50±1.2 & 51±1.5 & \textit{57±0.7} \\ 
\hline
\rowcolor[rgb]{0.906,0.976,0.937} ToeSegmentation2 & 63±1.8 & 63±1.8 & 63±1.8 & 55±5.5 & 63±1.8 & \textit{67±3} \\ 
\hline
Trace & 29±2.7 & 29±2.4 & 29±2.4 & 29±2.4 & 29±2.9 & \textit{89±1.8} \\ 
\hline
\rowcolor[rgb]{0.906,0.976,0.937} TwoLeadECG & 45±2.2 & 44±2.3 & 44±2.3 & 37±1.8 & 45±2.2 & \textit{77±0.7} \\ 
\hline
Two\_Patterns & 32±1.8 & 31±1.6 & 31±1.6 & 12±0.2 & 31±1.7 & \textit{96±0.4} \\ 
\hline
\rowcolor[rgb]{0.906,0.976,0.937} wafer & 39±1.5 & 39±1.5 & 39±1.5 & 21±1.5 & 39±1.5 & \textit{96±0.9} \\ 
\hline
Wine & 45±0 & 45±0 & 45±0 & 45±0 & 45±0 & \textit{56±0} \\ 
\hline
\rowcolor[rgb]{0.906,0.976,0.937} WordsSynonyms & 40±1.2 & 38±0.5 & 38±0.5 & 32±1 & 39±1.1 & \textit{53±0.4} \\ 
\hline
Worms & 28±0.4 & 27±0.9 & 27±0.9 & 24±1.5 & 28±0.6 & \textit{36±1.2} \\ 
\hline
\rowcolor[rgb]{0.906,0.976,0.937} WormsTwoClass & 49±1.2 & 49±1 & 49±1 & 47±1.4 & 49±1 & \textit{60±1} \\
\hline
\end{longtable}
\arrayrulecolor{black}

\arrayrulecolor[rgb]{0.8,0.8,0.8}
\begin{longtable}{|l|r|r|r|r|r|r|} 
\caption{Fully Convolutional Network (FCN) classification result on UCR repository under five adversarial attacks.}\label{app_tab:FCN}\\
\hline
\multicolumn{7}{|c|}{{\cellcolor[rgb]{0.545,0.765,0.29}}\textbf{FCN}} \endfirsthead 
\hline
\textbf{Datasets~~~~~~~~~~~~~~~~~~~~~~~~~~~~~} & \multicolumn{1}{c|}{\textbf{FGSM}} & \multicolumn{1}{c|}{\textbf{PGD}} & \multicolumn{1}{c|}{\textbf{BIM}} & \multicolumn{1}{c|}{\begin{tabular}[c]{@{}c@{}}\textbf{Carlini}\\\textbf{Wagner}\\\textbf{$\normltwo$}\end{tabular}} & \multicolumn{1}{c|}{\textbf{MIM}} & \multicolumn{1}{c|}{\begin{tabular}[c]{@{}c@{}}\textit{No}\\\textit{Attack}\end{tabular}} \\ 
\hline
\rowcolor[rgb]{0.933,0.969,0.89} 50words & 3±0.5 & 6±1.4 & 6±1.4 & 18±3.6 & 4±1.3 & \textit{29±16} \\ 
\hline
Adiac & 5±1.8 & 7±3.8 & 7±3.8 & 11±2.1 & 7±3.5 & \textit{24±17.7} \\ 
\hline
\rowcolor[rgb]{0.933,0.969,0.89} ArrowHead & 40±0 & 14±6.2 & 14±6.2 & 14±6.5 & 15±6 & \textit{80±6.6} \\ 
\hline
Beef & 26±10.2 & 23±9.7 & 23±9.7 & 23±12.7 & 22±7.7 & \textit{52±9.7} \\ 
\hline
\rowcolor[rgb]{0.933,0.969,0.89} BeetleFly & 50±0 & 20±5 & 20±5 & 20±5 & 20±5 & \textit{80±5} \\ 
\hline
BirdChicken & 50±0 & 15±10 & 15±10 & 7±2.9 & 22±2.9 & \textit{94±2.9} \\ 
\hline
\rowcolor[rgb]{0.933,0.969,0.89} Car & 22±0 & 40±27.5 & 40±27.5 & 40±26.2 & 40±25.1 & \textit{47±23.4} \\ 
\hline
CBF & 83±1.2 & 79±1.6 & 79±1.6 & 1±0.1 & 81±1.3 & \textit{100±0.2} \\ 
\hline
\rowcolor[rgb]{0.933,0.969,0.89} \begin{tabular}[c]{@{}>{\cellcolor[rgb]{0.933,0.969,0.89}}l@{}}Chlorine\\Concentration\end{tabular} & 39±19.5 & 39±19.8 & 39±19.8 & 38±19.1 & 39±19.8 & \textit{54±18.5} \\ 
\hline
Coffee & 0±0 & 0±0 & 0±0 & 0±0 & 0±0 & \textit{100±0} \\ 
\hline
\rowcolor[rgb]{0.933,0.969,0.89} Computers & 44±10 & 19±5.7 & 19±5.7 & 16±6.1 & 28±11 & \textit{85±6.1} \\ 
\hline
Cricket\_X & 16±5.7 & 11±1.8 & 11±1.8 & 13±2.3 & 11±3 & \textit{72±3.7} \\ 
\hline
\rowcolor[rgb]{0.933,0.969,0.89} Cricket\_Y & 19±1.9 & 16±3.1 & 16±3.1 & 16±2.9 & 16±3.3 & \textit{69±7.5} \\ 
\hline
Cricket\_Z & 13±1.1 & 11±3.2 & 11±3.2 & 14±3.5 & 11±2.1 & \textit{72±5.1} \\ 
\hline
\rowcolor[rgb]{0.933,0.969,0.89} \begin{tabular}[c]{@{}>{\cellcolor[rgb]{0.933,0.969,0.89}}l@{}}DiatomSize\\Reduction\end{tabular} & 16±4.9 & 6±0.9 & 6±0.9 & 7±0.5 & 7±0.7 & \textit{93±0.7} \\ 
\hline
\begin{tabular}[c]{@{}l@{}}DistalPhalanx\\OutlineAgeGroup\end{tabular} & 19±4.7 & 19±4.4 & 19±4.4 & 19±4.4 & 19±4.4 & \textit{80±4.3} \\ 
\hline
\rowcolor[rgb]{0.933,0.969,0.89} \begin{tabular}[c]{@{}>{\cellcolor[rgb]{0.933,0.969,0.89}}l@{}}DistalPhalanx\\OutlineCorrect\end{tabular} & 38±9.6 & 32±6.1 & 32±6.1 & 32±6.2 & 33±6.6 & \textit{69±6.1} \\ 
\hline
\begin{tabular}[c]{@{}l@{}}Distal\\PhalanxTW\end{tabular} & 15±1.1 & 17±1.2 & 17±1.2 & 17±1.1 & 17±1.1 & \textit{73±2.1} \\ 
\hline
\rowcolor[rgb]{0.933,0.969,0.89} Earthquakes & 36±4.1 & 34±3.2 & 34±3.2 & 25±2.5 & 35±3.3 & \textit{76±2.5} \\ 
\hline
ECG200 & 49±6.5 & 16±3.1 & 16±3.1 & 11±1.8 & 24±5 & \textit{89±1.8} \\ 
\hline
\rowcolor[rgb]{0.933,0.969,0.89} ECG5000 & 69±6.9 & 33±24.7 & 33±24.7 & 4±0.4 & 51±12.5 & \textit{94±0.4} \\ 
\hline
ECGFiveDays & 38±9.5 & 2±0.2 & 2±0.2 & 2±0.3 & 2±0.3 & \textit{99±0.3} \\ 
\hline
\rowcolor[rgb]{0.933,0.969,0.89} ElectricDevices & 43±1.3 & 32±2.7 & 32±2.7 & 14±3.3 & 35±2.9 & \textit{70±3.7} \\ 
\hline
FaceAll & 66±0.7 & 41±0.4 & 41±0.4 & 8±2.7 & 57±0.4 & \textit{90±2.8} \\ 
\hline
\rowcolor[rgb]{0.933,0.969,0.89} FaceFour & 6±2.3 & 3±1.8 & 3±1.8 & 5±1.8 & 3±1.2 & \textit{94±0.7} \\ 
\hline
FacesUCR & 68±2.4 & 40±7.9 & 40±7.9 & 4±0.7 & 56±4.4 & \textit{93±0.8} \\ 
\hline
\rowcolor[rgb]{0.933,0.969,0.89} FISH & 13±0.4 & 19±11.5 & 19±11.5 & 22±11.9 & 18±11 & \textit{60±2.9} \\ 
\hline
Gun\_Point & 51±2.7 & 2±0.7 & 2±0.7 & 1±0.4 & 4±2.4 & \textit{100±0.4} \\ 
\hline
\rowcolor[rgb]{0.933,0.969,0.89} Ham & 37±3.4 & 37±3.5 & 37±3.5 & 37±3.5 & 37±3.5 & \textit{64±3.5} \\ 
\hline
Haptics & 23±3.1 & 18±4.8 & 18±4.8 & 19±5 & 18±4.8 & \textit{29±3.4} \\ 
\hline
\rowcolor[rgb]{0.933,0.969,0.89} Herring & 60±0 & 46±8.2 & 46±8.2 & 49±11.9 & 54±5.5 & \textit{60±0} \\ 
\hline
InlineSkate & 16±0.5 & 13±5.2 & 13±5.2 & 16±6.7 & 13±4.5 & \textit{22±7.6} \\ 
\hline
\rowcolor[rgb]{0.933,0.969,0.89} \begin{tabular}[c]{@{}>{\cellcolor[rgb]{0.933,0.969,0.89}}l@{}}InsectWingbeat\\Sound\end{tabular} & 13±1.8 & 11±1.3 & 11±1.3 & 12±1.5 & 11±1.4 & \textit{23±4.4} \\ 
\hline
\begin{tabular}[c]{@{}l@{}}ItalyPower\\Demand\end{tabular} & 84±1 & 81±1.7 & 81±1.7 & 5±0.5 & 83±1.5 & \textit{96±0.3} \\ 
\hline
\rowcolor[rgb]{0.933,0.969,0.89} \begin{tabular}[c]{@{}>{\cellcolor[rgb]{0.933,0.969,0.89}}l@{}}LargeKitchen\\Appliances\end{tabular} & 50±4.9 & 32±23.7 & 32±23.7 & 21±17.5 & 45±13.9 & \textit{74±16} \\ 
\hline
Lighting2 & 40±1.7 & 29±1 & 29±1 & 29±1 & 30±1.7 & \textit{72±1} \\ 
\hline
\rowcolor[rgb]{0.933,0.969,0.89} Lighting7 & 32±7.6 & 19±2.9 & 19±2.9 & 17±3.5 & 23±4.2 & \textit{74±1.6} \\ 
\hline
Meat & 34±0 & 45±13.7 & 45±13.7 & 52±24.9 & 47±11.7 & \textit{34±0} \\ 
\hline
\rowcolor[rgb]{0.933,0.969,0.89} MedicalImages & 23±6.8 & 14±2 & 14±2 & 14±3.1 & 16±1.2 & \textit{77±2.8} \\ 
\hline
\begin{tabular}[c]{@{}l@{}}MiddlePhalanx\\OutlineAgeGroup\end{tabular} & 18±6.6 & 18±5.9 & 18±5.9 & 17±5.7 & 18±6.1 & \textit{70±6.7} \\ 
\hline
\rowcolor[rgb]{0.933,0.969,0.89} \begin{tabular}[c]{@{}>{\cellcolor[rgb]{0.933,0.969,0.89}}l@{}}MiddlePhalanx\\OutlineCorrect\end{tabular} & 44±22.5 & 43±21.6 & 43±21.6 & 45±24.2 & 43±21.6 & \textit{58±21.4} \\ 
\hline
MiddlePhalanxTW & 20±10 & 23±11 & 23±11 & 21±9 & 23±10.7 & \textit{48±12.8} \\ 
\hline
\rowcolor[rgb]{0.933,0.969,0.89} MoteStrain & 80±1 & 78±1.2 & 78±1.2 & 10±0.5 & 79±1.5 & \textit{91±0.5} \\ 
\hline
OliveOil & 18±19.3 & 16±21.2 & 16±21.2 & 18±19.3 & 18±19.3 & \textit{56±15.1} \\ 
\hline
\rowcolor[rgb]{0.933,0.969,0.89} OSULeaf & 14±0 & 12±4 & 12±4 & 12±4.4 & 11±4.1 & \textit{75±16.7} \\ 
\hline
\begin{tabular}[c]{@{}l@{}}Phalanges\\OutlinesCorrect\end{tabular} & 36±2.5 & 36±2.5 & 36±2.5 & 36±2.6 & 36±2.5 & \textit{65±2.6} \\ 
\hline
\rowcolor[rgb]{0.933,0.969,0.89} Plane & 40±5.8 & 11±3.9 & 11±3.9 & 0±0 & 25±6.5 & \textit{100±0} \\ 
\hline
\begin{tabular}[c]{@{}l@{}}ProximalPhalanx\\OutlineAgeGroup\end{tabular} & 32±23.7 & 22±8.8 & 22±8.8 & 25±10.7 & 22±8.8 & \textit{64±18.9} \\ 
\hline
\rowcolor[rgb]{0.933,0.969,0.89} \begin{tabular}[c]{@{}>{\cellcolor[rgb]{0.933,0.969,0.89}}l@{}}ProximalPhalanx\\OutlineCorrect\end{tabular} & 32±26.8 & 31±26.4 & 31±26.4 & 31±26.2 & 31±26.8 & \textit{70±26.2} \\ 
\hline
\begin{tabular}[c]{@{}l@{}}Proximal\\PhalanxTW\end{tabular} & 18±8.2 & 14±3.1 & 14±3.1 & 15±4.7 & 14±2.9 & \textit{75±2.9} \\ 
\hline
\rowcolor[rgb]{0.933,0.969,0.89} \begin{tabular}[c]{@{}>{\cellcolor[rgb]{0.933,0.969,0.89}}l@{}}Refrigeration\\Devices\end{tabular} & 40±3.5 & 36±0.9 & 36±0.9 & 35±1.7 & 36±1 & \textit{46±1.7} \\ 
\hline
ScreenType & 33±3.3 & 28±3.6 & 28±3.6 & 27±3.6 & 29±4.3 & \textit{62±5.2} \\ 
\hline
\rowcolor[rgb]{0.933,0.969,0.89} ShapeletSim & 8±3.7 & 8±3.1 & 8±3.1 & 8±2.8 & 8±3.1 & \textit{93±2.8} \\ 
\hline
ShapesAll & 4±1.4 & 3±2.9 & 3±2.9 & 7±0.6 & 3±1.9 & \textit{19±18} \\ 
\hline
\rowcolor[rgb]{0.933,0.969,0.89} \begin{tabular}[c]{@{}>{\cellcolor[rgb]{0.933,0.969,0.89}}l@{}}SmallKitchen\\Appliances\end{tabular} & 53±16.7 & 37±18.1 & 37±18.1 & 39±22.6 & 41±11.1 & \textit{43±12.3} \\ 
\hline
\begin{tabular}[c]{@{}l@{}}SonyAIBO\\RobotSurface\end{tabular} & 84±2.2 & 82±2.7 & 82±2.7 & 5±0.3 & 83±2.7 & \textit{97±0.6} \\ 
\hline
\rowcolor[rgb]{0.933,0.969,0.89} \begin{tabular}[c]{@{}>{\cellcolor[rgb]{0.933,0.969,0.89}}l@{}}SonyAIBO\\RobotSurfaceII\end{tabular} & 86±1.5 & 84±2.1 & 84±2.1 & 3±0.5 & 85±1.7 & \textit{98±0.5} \\ 
\hline
Strawberry & 44±20.8 & 31±8.8 & 31±8.8 & 31±8.9 & 31±9.1 & \textit{70±8.8} \\ 
\hline
\rowcolor[rgb]{0.933,0.969,0.89} SwedishLeaf & 28±1.7 & 10±2.6 & 10±2.6 & 6±3.6 & 13±3.3 & \textit{93±3.6} \\ 
\hline
Symbols & 36±3.2 & 6±1.6 & 6±1.6 & 5±0.6 & 15±1.9 & \textit{94±1.3} \\ 
\hline
\rowcolor[rgb]{0.933,0.969,0.89} synthetic\_control & 95±1 & 95±1.3 & 95±1.3 & 3±0.9 & 95±1.2 & \textit{98±0.7} \\ 
\hline
ToeSegmentation1 & 41±6.2 & 11±0.8 & 11±0.8 & 3±0.7 & 18±3 & \textit{98±0.7} \\ 
\hline
\rowcolor[rgb]{0.933,0.969,0.89} ToeSegmentation2 & 43±1.4 & 26±2.3 & 26±2.3 & 14±2.8 & 36±0.5 & \textit{87±2.8} \\ 
\hline
Trace & 52±18.6 & 18±8.9 & 18±8.9 & 1±0.6 & 43±2.9 & \textit{100±0.6} \\ 
\hline
\rowcolor[rgb]{0.933,0.969,0.89} TwoLeadECG & 7±3.1 & 2±0.4 & 2±0.4 & 1±0.1 & 3±0.7 & \textit{100±0.1} \\ 
\hline
Two\_Patterns & 34±7.3 & 15±0.7 & 15±0.7 & 15±0.7 & 19±2.3 & \textit{86±0.7} \\ 
\hline
\rowcolor[rgb]{0.933,0.969,0.89} wafer & 8±3.2 & 3±0.9 & 3±0.9 & 1±0.2 & 3±1.3 & \textit{100±0.2} \\ 
\hline
Wine & 50±0 & 50±0 & 50±0 & 50±0 & 50±0 & \textit{50±0} \\ 
\hline
\rowcolor[rgb]{0.933,0.969,0.89} WordsSynonyms & 5±2.2 & 9±3.3 & 9±3.3 & 12±1.5 & 6±1.9 & \textit{30±10.2} \\ 
\hline
Worms & 17±1.7 & 21±3.6 & 21±3.6 & 21±5.3 & 21±3.4 & \textit{48±7.3} \\ 
\hline
\rowcolor[rgb]{0.933,0.969,0.89} WormsTwoClass & 48±5 & 39±2.3 & 39±2.3 & 39±2.5 & 40±4.2 & \textit{62±2.3} \\
\hline
\end{longtable}
\arrayrulecolor{black}

\arrayrulecolor[rgb]{0.8,0.8,0.8}
\begin{longtable}{|l|r|r|r|r|r|r|} 
\caption{ResNet classification result on UCR repository under five adversarial attacks.}\label{app_tab:ResNet}\\
\hline
\multicolumn{7}{|c|}{{\cellcolor[rgb]{0.8,0.651,0.467}}\textbf{ResNet}} \endfirsthead 
\hline
\textbf{Datasets~~~~~~~~~~~~~~~~~~~~~~~~~~~~~~} & \multicolumn{1}{c|}{\textbf{FGSM}} & \multicolumn{1}{c|}{\textbf{PGD}} & \multicolumn{1}{c|}{\textbf{BIM}} & \multicolumn{1}{c|}{\begin{tabular}[c]{@{}c@{}}\textbf{Carlini}\\\textbf{Wagner}\\\textbf{$\normltwo$}\end{tabular}} & \multicolumn{1}{c|}{\textbf{MIM}} & \multicolumn{1}{c|}{\begin{tabular}[c]{@{}c@{}}\textit{No}\\\textit{Attack}\end{tabular}} \\ 
\hline
\rowcolor[rgb]{0.973,0.949,0.922} 50words & 8±2.3 & 10±1 & 10±1 & 13±1.5 & 9±1.5 & \textit{67±0.7} \\ 
\hline
Adiac & 5±0.2 & 10±1.2 & 10±1.2 & 10±0.2 & 10±0.4 & \textit{82±0.7} \\ 
\hline
\rowcolor[rgb]{0.973,0.949,0.922} ArrowHead & 34±11.5 & 13±0.9 & 13±0.9 & 13±1.5 & 15±1 & \textit{79±2.3} \\ 
\hline
Beef & 24±8.9 & 19±5.1 & 19±5.1 & 18±3.9 & 22±3.9 & \textit{74±3.4} \\ 
\hline
\rowcolor[rgb]{0.973,0.949,0.922} BeetleFly & 29±5.8 & 17±5.8 & 17±5.8 & 17±5.8 & 17±5.8 & \textit{84±5.8} \\ 
\hline
BirdChicken & 54±5.8 & 14±2.9 & 14±2.9 & 14±2.9 & 20±5 & \textit{87±2.9} \\ 
\hline
\rowcolor[rgb]{0.973,0.949,0.922} Car & 20±1 & 9±4.5 & 9±4.5 & 8±3.9 & 10±4.9 & \textit{89±3.5} \\ 
\hline
CBF & 89±1.4 & 87±1.8 & 87±1.8 & 1±0.2 & 88±1.6 & \textit{100±0.2} \\ 
\hline
\rowcolor[rgb]{0.973,0.949,0.922} \begin{tabular}[c]{@{}>{\cellcolor[rgb]{0.973,0.949,0.922}}l@{}}Chlorine\\Concentration\end{tabular} & 14±0.4 & 14±0.8 & 14±0.8 & 13±0.4 & 14±0.7 & \textit{82±1.1} \\ 
\hline
Coffee & 0±0 & 0±0 & 0±0 & 0±0 & 0±0 & \textit{100±0} \\ 
\hline
\rowcolor[rgb]{0.973,0.949,0.922} Computers & 58±5.4 & 24±1.3 & 24±1.3 & 20±3.2 & 45±5.1 & \textit{82±2.6} \\ 
\hline
Cricket\_X & 33±3 & 17±2.5 & 17±2.5 & 14±2.1 & 27±1.9 & \textit{76±2.4} \\ 
\hline
\rowcolor[rgb]{0.973,0.949,0.922} Cricket\_Y & 23±0.6 & 13±0.7 & 13±0.7 & 13±0.6 & 16±1.7 & \textit{80±1.1} \\ 
\hline
Cricket\_Z & 28±2.9 & 14±2 & 14±2 & 13±0.8 & 22±2.4 & \textit{78±1.4} \\ 
\hline
\rowcolor[rgb]{0.973,0.949,0.922} \begin{tabular}[c]{@{}>{\cellcolor[rgb]{0.973,0.949,0.922}}l@{}}DiatomSize\\Reduction\end{tabular} & 10±4.1 & 4±1.5 & 4±1.5 & 5±2 & 4±1.4 & \textit{97±1.9} \\ 
\hline
\begin{tabular}[c]{@{}l@{}}DistalPhalanx\\OutlineAgeGroup\end{tabular} & 18±2.4 & 17±1.8 & 17±1.8 & 17±2 & 17±1.8 & \textit{81±1.8} \\ 
\hline
\rowcolor[rgb]{0.973,0.949,0.922} \begin{tabular}[c]{@{}>{\cellcolor[rgb]{0.973,0.949,0.922}}l@{}}DistalPhalanx\\OutlineCorrect\end{tabular} & 29±3.6 & 23±1 & 23±1 & 21±1.2 & 25±1.7 & \textit{80±1} \\ 
\hline
\begin{tabular}[c]{@{}l@{}}Distal\\PhalanxTW\end{tabular} & 15±0.3 & 15±0.8 & 15±0.8 & 14±0.6 & 15±0.9 & \textit{76±0.7} \\ 
\hline
\rowcolor[rgb]{0.973,0.949,0.922} Earthquakes & 48±2.9 & 45±2.7 & 45±2.7 & 24±1 & 46±3.1 & \textit{80±1.2} \\ 
\hline
ECG200 & 69±4.4 & 50±11.6 & 50±11.6 & 13±2.1 & 63±4.1 & \textit{88±2.4} \\ 
\hline
\rowcolor[rgb]{0.973,0.949,0.922} ECG5000 & 73±0.8 & 61±1.3 & 61±1.3 & 5±0.3 & 66±1.3 & \textit{94±0.3} \\ 
\hline
ECGFiveDays & 33±16.2 & 4±1.6 & 4±1.6 & 3±0.6 & 6±3.8 & \textit{98±0.7} \\ 
\hline
\rowcolor[rgb]{0.973,0.949,0.922} ElectricDevices & 41±2.1 & 31±1.7 & 31±1.7 & 15±2.4 & 36±2.2 & \textit{70±4.5} \\ 
\hline
FaceAll & 76±0.4 & 69±1 & 69±1 & 11±0.5 & 74±0.7 & \textit{83±1.6} \\ 
\hline
\rowcolor[rgb]{0.973,0.949,0.922} FaceFour & 30±5.2 & 9±2.4 & 9±2.4 & 4±2.9 & 22±3.5 & \textit{95±0.7} \\ 
\hline
FacesUCR & 74±1.4 & 64±2.3 & 64±2.3 & 3±0.8 & 70±1.6 & \textit{95±0.4} \\ 
\hline
\rowcolor[rgb]{0.973,0.949,0.922} FISH & 13±0 & 3±0.9 & 3±0.9 & 3±1.2 & 3±0.9 & \textit{98±1} \\ 
\hline
Gun\_Point & 23±5.6 & 6±2 & 6±2 & 1±0.4 & 10±0.7 & \textit{100±0} \\ 
\hline
\rowcolor[rgb]{0.973,0.949,0.922} Ham & 30±2.9 & 29±2 & 29±2 & 30±2.4 & 29±2 & \textit{72±2} \\ 
\hline
Haptics & 20±0.2 & 22±3.1 & 22±3.1 & 21±3.7 & 21±3.8 & \textit{49±4} \\ 
\hline
\rowcolor[rgb]{0.973,0.949,0.922} Herring & 49±11 & 41±1 & 41±1 & 41±1 & 41±1 & \textit{60±1} \\ 
\hline
InlineSkate & 15±1.3 & 19±2 & 19±2 & 19±2.9 & 19±1.9 & \textit{32±3.1} \\ 
\hline
\rowcolor[rgb]{0.973,0.949,0.922} \begin{tabular}[c]{@{}>{\cellcolor[rgb]{0.973,0.949,0.922}}l@{}}InsectWingbeat\\Sound\end{tabular} & 22±0.5 & 23±0.6 & 23±0.6 & 23±0.4 & 24±0.3 & \textit{46±1.1} \\ 
\hline
\begin{tabular}[c]{@{}l@{}}ItalyPower\\Demand\end{tabular} & 87±1.3 & 86±0.8 & 86±0.8 & 7±0.9 & 86±1.3 & \textit{97±0.2} \\ 
\hline
\rowcolor[rgb]{0.973,0.949,0.922} \begin{tabular}[c]{@{}>{\cellcolor[rgb]{0.973,0.949,0.922}}l@{}}LargeKitchen\\Appliances\end{tabular} & 59±2.8 & 32±2.7 & 32±2.7 & 8±1.4 & 47±1.2 & \textit{90±0.8} \\ 
\hline
Lighting2 & 46±0 & 42±2.6 & 42±2.6 & 27±1.7 & 43±1.7 & \textit{74±1.7} \\ 
\hline
\rowcolor[rgb]{0.973,0.949,0.922} Lighting7 & 36±3.7 & 20±4.2 & 20±4.2 & 19±2.1 & 24±7.7 & \textit{74±4.2} \\ 
\hline
Meat & 17±15.5 & 8±5.4 & 8±5.4 & 8±5.4 & 8±5.4 & \textit{93±5.4} \\ 
\hline
\rowcolor[rgb]{0.973,0.949,0.922} MedicalImages & 47±5 & 28±3.8 & 28±3.8 & 15±2.5 & 36±2.4 & \textit{78±0.7} \\ 
\hline
\begin{tabular}[c]{@{}l@{}}MiddlePhalanx\\OutlineAgeGroup\end{tabular} & 16±1.5 & 16±0.7 & 16±0.7 & 15±0.2 & 16±0.8 & \textit{75±1} \\ 
\hline
\rowcolor[rgb]{0.973,0.949,0.922} \begin{tabular}[c]{@{}>{\cellcolor[rgb]{0.973,0.949,0.922}}l@{}}MiddlePhalanx\\OutlineCorrect\end{tabular} & 27±9.1 & 27±9 & 27±9 & 27±9.1 & 27±9 & \textit{74±9.2} \\ 
\hline
\begin{tabular}[c]{@{}l@{}}Middle\\PhalanxTW\end{tabular} & 15±2.8 & 17±0.4 & 17±0.4 & 17±0.6 & 17±0.7 & \textit{62±0.8} \\ 
\hline
\rowcolor[rgb]{0.973,0.949,0.922} MoteStrain & 76±0.9 & 73±1.1 & 73±1.1 & 10±0.8 & 75±1.1 & \textit{91±0.8} \\ 
\hline
OliveOil & 14±0 & 17±5.8 & 17±5.8 & 18±3.9 & 17±5.8 & \textit{79±2} \\ 
\hline
\rowcolor[rgb]{0.973,0.949,0.922} OSULeaf & 14±1 & 6±2.2 & 6±2.2 & 5±1.9 & 6±2.2 & \textit{94±2.8} \\ 
\hline
\begin{tabular}[c]{@{}l@{}}Phalanges\\OutlinesCorrect\end{tabular} & 27±3 & 17±0.9 & 17±0.9 & 18±0.7 & 17±0.9 & \textit{84±0.9} \\ 
\hline
\rowcolor[rgb]{0.973,0.949,0.922} Plane & 73±6.2 & 41±6.4 & 41±6.4 & 0±0 & 63±5.3 & \textit{100±0} \\ 
\hline
\begin{tabular}[c]{@{}l@{}}ProximalPhalanx\\OutlineAgeGroup\end{tabular} & 16±4.8 & 15±0.8 & 15±0.8 & 16±1.5 & 15±0.8 & \textit{86±0.6} \\ 
\hline
\rowcolor[rgb]{0.973,0.949,0.922} \begin{tabular}[c]{@{}>{\cellcolor[rgb]{0.973,0.949,0.922}}l@{}}ProximalPhalanx\\OutlineCorrect\end{tabular} & 16±2.6 & 11±1.6 & 11±1.6 & 11±1.7 & 11±1.6 & \textit{90±1.6} \\ 
\hline
\begin{tabular}[c]{@{}l@{}}Proximal\\PhalanxTW\end{tabular} & 8±1.2 & 13±0.5 & 13±0.5 & 14±0.3 & 14±0.4 & \textit{82±0.5} \\ 
\hline
\rowcolor[rgb]{0.973,0.949,0.922} \begin{tabular}[c]{@{}>{\cellcolor[rgb]{0.973,0.949,0.922}}l@{}}Refrigeration\\Devices\end{tabular} & 35±2.5 & 34±3.1 & 34±3.1 & 31±2.3 & 34±3.1 & \textit{54±0.6} \\ 
\hline
ScreenType & 35±7 & 29±2.6 & 29±2.6 & 28±3.5 & 32±4.5 & \textit{61±3.8} \\ 
\hline
\rowcolor[rgb]{0.973,0.949,0.922} ShapeletSim & 13±7.9 & 12±8.6 & 12±8.6 & 10±10.2 & 13±8.1 & \textit{91±9.9} \\ 
\hline
ShapesAll & 7±0.7 & 3±0.3 & 3±0.3 & 5±0.3 & 4±0.7 & \textit{88±0.5} \\ 
\hline
\rowcolor[rgb]{0.973,0.949,0.922} \begin{tabular}[c]{@{}>{\cellcolor[rgb]{0.973,0.949,0.922}}l@{}}SmallKitchen\\Appliances\end{tabular} & 44±4.5 & 28±5.6 & 28±5.6 & 29±7.7 & 34±5.2 & \textit{56±16} \\ 
\hline
\begin{tabular}[c]{@{}l@{}}SonyAIBO\\RobotSurface\end{tabular} & 80±2.3 & 79±2.9 & 79±2.9 & 14±3.2 & 79±2.5 & \textit{92±0.9} \\ 
\hline
\rowcolor[rgb]{0.973,0.949,0.922} \begin{tabular}[c]{@{}>{\cellcolor[rgb]{0.973,0.949,0.922}}l@{}}SonyAIBO\\RobotSurfaceII\end{tabular} & 81±1.1 & 79±1.6 & 79±1.6 & 4±0.8 & 80±1 & \textit{98±0.8} \\ 
\hline
Strawberry & 24±16 & 22±17.7 & 22±17.7 & 22±17.6 & 22±17.7 & \textit{80±17.6} \\ 
\hline
\rowcolor[rgb]{0.973,0.949,0.922} SwedishLeaf & 34±0.8 & 16±0.5 & 16±0.5 & 4±0.5 & 22±0.9 & \textit{96±0.4} \\ 
\hline
Symbols & 32±2.1 & 8±0.5 & 8±0.5 & 5±1.6 & 16±1.5 & \textit{95±1.7} \\ 
\hline
\rowcolor[rgb]{0.973,0.949,0.922} synthetic\_control & 95±0.7 & 95±0.4 & 95±0.4 & 20±4 & 95±0.7 & \textit{100±0.4} \\ 
\hline
ToeSegmentation1 & 54±1.8 & 31±2.5 & 31±2.5 & 4±0.7 & 39±2 & \textit{97±0.7} \\ 
\hline
\rowcolor[rgb]{0.973,0.949,0.922} ToeSegmentation2 & 45±5.2 & 35±5.9 & 35±5.9 & 11±2.5 & 41±4.3 & \textit{90±2.5} \\ 
\hline
Trace & 30±2.1 & 13±9.7 & 13±9.7 & 2±1.6 & 37±8.6 & \textit{98±0} \\ 
\hline
\rowcolor[rgb]{0.973,0.949,0.922} TwoLeadECG & 8±4.8 & 2±0.6 & 2±0.6 & 1±0.5 & 4±1.7 & \textit{100±0.3} \\ 
\hline
Two\_Patterns & 68±1.9 & 42±6.2 & 42±6.2 & 6±1.1 & 56±3.8 & \textit{96±1} \\ 
\hline
\rowcolor[rgb]{0.973,0.949,0.922} wafer & 17±11.8 & 7±7.8 & 7±7.8 & 2±0.2 & 11±10.8 & \textit{100±0.1} \\ 
\hline
Wine & 34±16 & 25±8.4 & 25±8.4 & 25±8.4 & 25±8.4 & \textit{76±8.4} \\ 
\hline
\rowcolor[rgb]{0.973,0.949,0.922} WordsSynonyms & 15±3.1 & 14±1 & 14±1 & 16±0.4 & 14±1.4 & \textit{54±1.3} \\ 
\hline
Worms & 26±2 & 21±1.5 & 21±1.5 & 19±0.9 & 25±0.4 & \textit{63±2} \\ 
\hline
\rowcolor[rgb]{0.973,0.949,0.922} WormsTwoClass & 54±2.7 & 29±2 & 29±2 & 27±2 & 32±1.4 & \textit{75±1.4} \\
\hline
\end{longtable}
\arrayrulecolor{black}